\documentclass[final]{cvpr}

\usepackage{times}
\usepackage{epsfig}
\usepackage{graphicx}
\usepackage{amsmath}
\usepackage{amssymb}
\usepackage{capt-of}

\usepackage{color}
\usepackage[table]{xcolor}
\usepackage{multirow}
\usepackage{tabularx}
\usepackage{hhline}
\usepackage{booktabs}
\usepackage{wrapfig}

\usepackage[accsupp]{axessibility}  %

\usepackage{xspace}

\usepackage{soul}

\definecolor{col1}{RGB}{232, 161, 148}
\definecolor{col2}{RGB}{148, 187, 232}
\definecolor{lightblue}{RGB}{225, 225, 255}
\definecolor{ms_note}{RGB}{0, 181, 190}

\renewcommand*{\ie}{i.e.\@\xspace}
\renewcommand*{\eg}{e.g.\@\xspace}

\newcommand*{\ea}{et al.\@\xspace}

\renewcommand{\paragraph}[1]{\vspace{2pt}\noindent\textbf{#1}\hspace{5pt}}

\usepackage[pagebackref=true,breaklinks=true,colorlinks,bookmarks=false]{hyperref}
\def\cvprPaperID{8061} %
\def\confYear{CVPR 2023}

\begin{document}

\makeatletter
\@namedef{ver@everyshi.sty}{}
\makeatother

\title{Virtual Occlusions Through Implicit Depth}

\author{Jamie Watson$^{1,3}$\hspace{15pt}Mohamed Sayed$^{1,3}$\hspace{15pt}Zawar Qureshi$^{1}$\hspace{15pt}Gabriel J.~Brostow$^{1,3}$\\Sara Vicente$^{1}$\hspace{15pt}Oisin Mac Aodha$^{2}$\hspace{15pt}Michael Firman$^{1}$\\$^{1}$Niantic \hspace{30pt} $^{2}$University of Edinburgh \hspace{30pt}$^{3}$UCL\vspace{1pt}\\ 
\small 
\href{https://nianticlabs.github.io/implicit-depth}{\texttt{https://nianticlabs.github.io/implicit-depth}}
}

\maketitle

\begin{abstract}
    For augmented reality (AR), it is important that virtual assets appear to `sit among' real world objects. The virtual element should variously occlude and be occluded by real matter, based on a plausible depth ordering. This occlusion should be consistent over time as the viewer's camera moves. Unfortunately, small mistakes in the estimated scene depth can ruin the downstream occlusion mask, and thereby the AR illusion. Especially in real-time settings, depths inferred near boundaries or across time can be inconsistent. In this paper, we challenge the need for depth-regression as an intermediate step. 
    
    We instead propose an implicit model for depth and use that to predict the occlusion mask directly. The inputs to our network are one or more color images, plus the known depths of any virtual geometry. We show how our occlusion predictions are more accurate and more temporally stable than predictions derived from traditional depth-estimation models. We obtain state-of-the-art occlusion results on the challenging ScanNetv2 dataset and superior qualitative results on real scenes.

\end{abstract}

\section{Introduction}

Augmented reality and digital image editing usually entail compositing virtual rendered objects to look as if they are present in a real-world scene.
A key and elusive part of making this effect realistic is \emph{occlusion}.
Looking from a camera's perspective, a virtual object should appear partially hidden when part of it passes behind a real world object. 
In practice this comes down to estimating, for each pixel, if the final rendering pipeline should display the real world object there \vs showing the virtual object~\cite{lepetit2000semi,fuchs1998augmented,holynski2018fast}. 

\begin{figure}
    \centering
    \includegraphics[width=1.0\columnwidth]{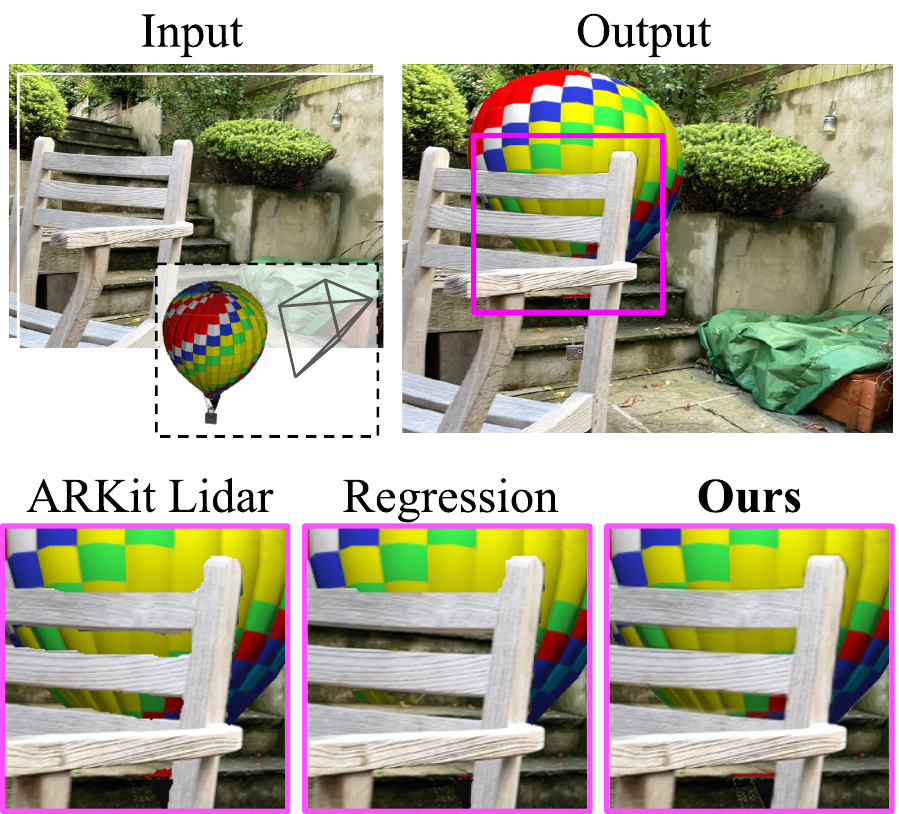}
    \caption{We address the problem of automatically estimating occlusion masks to realistically place virtual objects in real scenes. 
    Our approach, where we directly predict masks, leads to more accurate compositing compared with Lidar-based sensors or traditional state-of-the-art depth regression methods.}
    \label{fig:teaser}
    \vspace{-7pt}
\end{figure}

Typically, this per-pixel decision is approached by first estimating the depth of each pixel in the real world image \cite{du2020depthlab,valentin2018depth,holynski2018fast}. Obtaining the depth to each pixel on the \emph{virtual} object is trivial, and can be computed via traditional graphics pipelines \cite{hearn2004computer}. The final mask can be estimated by comparing the two depth maps: where the real world depth value is smaller, the virtual content is occluded, \ie masked.%

We propose an alternative, novel approach. Given images of the real world scene and the depth map of the virtual assets, our network \emph{directly estimates} the mask for compositing.
The key advantage is that the network no longer has to estimate the real-valued depth for every pixel, and instead focuses on the binary decision: is the virtual pixel in front or behind the real scene here?
Further, at inference time we can use the soft output of a sigmoid layer to softly blend between the real and virtual, which can give visually pleasing compositions \cite{jampani2021slide}, compared with those created by hard thresholding of depth maps (see Figure~\ref{fig:teaser}). 
Finally, temporal consistency can be improved using ideas from temporal semantic segmentation that were difficult to apply when regressing depth as an intermediate step.

\noindent We have three contributions: 
\vspace{-5pt}
\begin{enumerate}
    \itemsep-0.3em 
    \item We frame the problem of compositing a virtual object at a known depth into a real scene as a binary mask estimation problem. This is in contrast to previous approaches that estimate depth to solve this problem.
    \item We introduce metrics for occlusion evaluation, and find our method results in more accurate and visually pleasing composites compared with alternatives.
    \item We show that competing approaches flicker, leading to jarring occlusions. 
    By framing the problem as segmentation, we can use temporal smoothing methods, which result in smoother predictions compared to baselines.
\end{enumerate}
\vspace{-5pt}

Our `implicit depth' model ultimately results in state-of-the-art occlusions on the challenging ScanNetv2 dataset~\cite{dai2017scannet}.
Further, if depths are needed too, we can compute dense depth by gathering multiple binary masks. 
Surprisingly, this results in state-of-the-art depth estimation.

\section{Related work}

Early approaches to occluding virtual assets in real scenes relied on user annotations of object boundaries \cite{lepetit2000semi,ong1998resolving}, precluding general real-time use.
The typical approach for automated occlusion is to pixel-wise compare a depth map of the real scene  with the virtual depth map~\cite{breen1996interactive}.
The real image depth map can be estimated, \eg using structured light~\cite{fuchs1998augmented}, or from images~\cite{du2020depthlab,valentin2018depth}. 
When sparse depths are available, they can be densified~\cite{holynski2018fast} or used to rescale relative depths to metric depth~\cite{aleotti2020real}.
Direct estimation of an occlusion mask has been previously performed for segmentation for AR sky replacement~\cite{zou2022castle} or hands grabbing virtual objects~\cite{tang2020grabar}.
In contrast, our method enables general object compositing.
A detailed review of occlusion handling in AR is outlined in \cite{macedo2021occlusion,cao2021mobile}.

\paragraph{Depth estimation.}
Depth estimation is a key component of many AR occlusion systems.
Depth can be estimated directly if binocular cameras are available at test time~\cite{hirschmuller2007stereo,kendall2017end, chang2018pyramid,cheng2019learning}. %
This approach requires specialized hardware, as does depth estimation from structured light, Lidar, or time-of-flight devices \cite{giancola2018survey}.
Further, depth from such devices may not be accurate enough for realistic occlusions without further processing \cite{walton2017accurate,jorge2019dynamic}.

It is attractive to estimate depth directly from color images, for example from a \emph{single} image \cite{garg2016unsupervised,eigen2014depth,eigen2015predicting,godard2017unsupervised,xie2016deep3d}.
When a sequence of images is available, Multi-View Stereo (MVS) estimates depth for a reference image using one or more source images \cite{furukawa2015multi,schoenberger2016mvs}, which assumes that the scene being observed is static.
Recent MVS approaches match image pixels~\cite{wang2018mvdepthnet,huang2018DeepMVS} or deep features \cite{im2019dpsnet, duzceker2021deepvideomvs} to create a cost volume, which can then be processed using convolutional layers. 
Other works have improved upon this basic setup by refining the final output \cite{yao2018mvsnet}, through injection of additional metadata \cite{sayed2022simplerecon}, by handling occluded pixels between views~\cite{long2020occlusion}, or with a Gaussian process prior~\cite{hou2019multi}. 
Alternative approaches have dropped the reliance on supervised data through the use of self-supervision~\cite{godard2017unsupervised, garg2016unsupervised, godard2019digging, watson2021multidepth}.
There have been attempts to improve the quality of depth estimation around depth discontinuities, \eg using a gradient or normal loss \cite{Ranftl2020, li2018megadepth,hu2019revisiting} or a learned network to post-process predictions \cite{ramamonjisoa2020predicting}.
Higher quality depths can also be computed via an offline optimization on test sequences \cite{kopf2021robust,luo2020consistent,casser2018depth,chen2019self,kuznietsov2021comoda}, but this precludes online use.

In contrast to these depth estimation approaches, we \emph{directly} estimate an occlusion compositing mask.
However, in Section~\ref{sec:experiments} we show that our models can also be adapted to predict depths equivalent to state-of-the-art methods without requiring retraining.

\paragraph{Depth via classification.} 
Our implicit depth approach is related to classification-based depth estimation. 
Xie~\ea~\cite{xie2016deep3d} is an early approach which learned depth via classification, in the context of depth from stereo.
In \cite{fu2018deep}, the output domain is divided into discrete bins, and the final output head classifies each pixel as in front or behind each depth bin.
Alternatively, \cite{liu2019neural,cao2017estimating,yang2019inferring,tucker2020single} classify the probability that the depth lands in a bin itself. 
Other works have relaxed the requirement for fixed bin centres, allowing them to be adapted on a per-image \cite{bhatCVPR2021adabins} or per-pixel \cite{bhatECCV2022localbins} basis.
Bi3D~\cite{badki2020bi3d} pose stereo depth estimation as a binary classification task, but where a scalar query depth is provided to the network.
Also related to depth classification approaches are works which decompose images into two or more layers, \eg foreground and background~\cite{dhamo2018peeking,tulsiani2018layer,li2021mine}.

We also frame geometry estimation as classification, but our approach classifies if a per-pixel virtual depth is in front or behind the real world object at that depth.
We can therefore estimate a full compositing mask with a single forward pass, without a dense output tensor. 
We compare to classification approaches and show that our method is superior in terms of accuracy and compute.

\paragraph{Image and video segmentation.}
Our work is related to object segmentation \cite{he2017mask,zhang2021refinemask}, salient object segmentation \cite{jiang2011automatic}, and alpha-matting \cite{lin2021real,sengupta2020matting, chen2022pp}.
Like these, we estimate a binary mask, but our mask is conditioned on an input rendered virtual depth map.
Similar to video segmentation \cite{gadde2017semantic, perazzi2017learning}, we encourage temporal consistency across frames to prevent flicker.

\paragraph{Occlusion boundaries and regions.} 
There are works which focus on detecting pixels which become occluded and disoccluded between frames in a video~\cite{humayun_CVPR_2011_occlusions, wang2020occlusion, hoiem2007recovering, stein2009occlusion,sundberg2011occlusion}.
We differ as we recover the occlusion mask of a virtual object in real input images.

\paragraph{Implicit volumes.}
Finally, our approach is related to works on implicit volumes~\eg \cite{liu2019learning,park2019deepsdf,mescheder2019occupancy,runz2020frodo,saito2019pifu,saito2020pifuhd,chibane2020implicit,roddick2021road,peng2020convolutional}, where a 3D shape is represented by a trained multi-layer perceptron (MLP). When evaluated at each location in 3D space, the MLP's binary output indicates if that location is inside or outside an object.
However, we operate in image space, and predict if a pixel in the real world scene is in front or behind a virtual target's depth map.
Zhu~\ea~\cite{zhu2021rgb} used an implicit function for RGBD \emph{completion}, operating on ray-voxel pairs.
Neural radiance fields~\cite{mildenhall2020nerf,barron2021mipnerf,Niemeyer2021Regnerf,yu2020pixelnerf} (NeRFs) are an alternative implicit approach which can estimate depths and color images from novel viewpoints.
This can be used for AR effects, but NeRFs are typically not suited to online applications in novel scenes.

\section{Method}

\begin{figure*}
    \centering
    \includegraphics[width=0.99\textwidth]{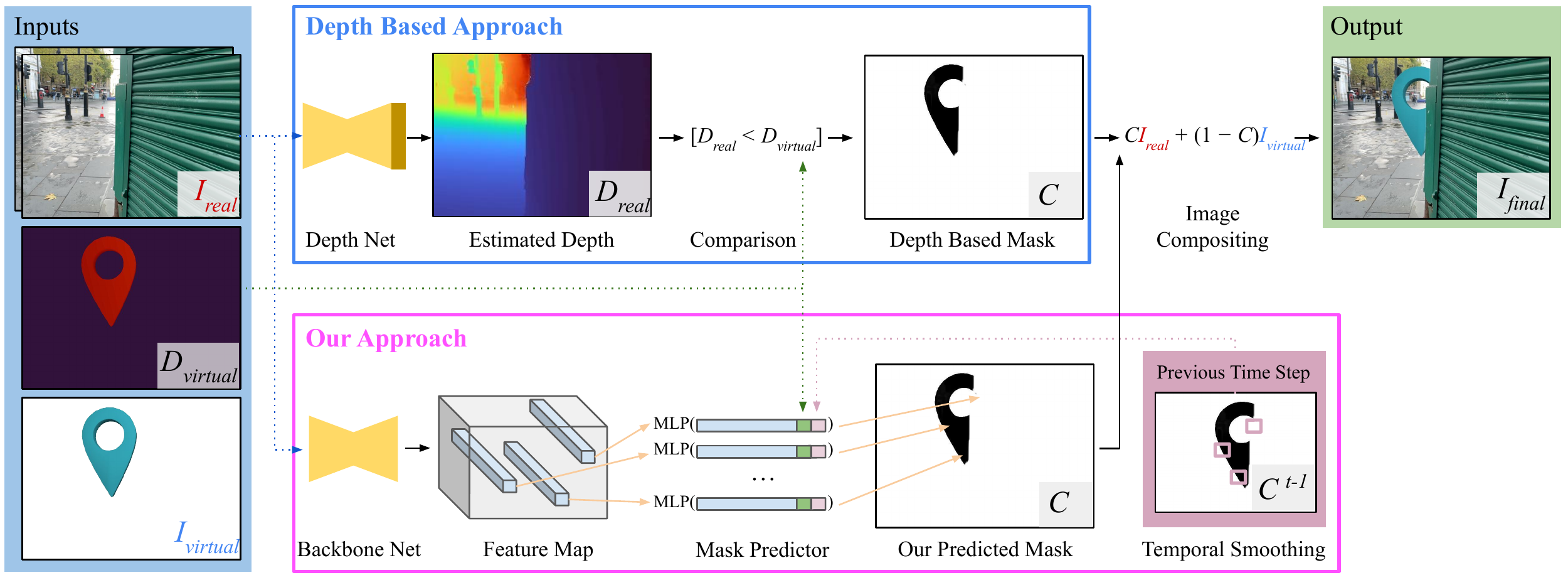}
    \caption{Left: Given RGB images of a real scene, and renderings of a virtual asset, our aim is to realistically composite the virtual asset into the scene. Top: Conventional approaches first estimate a depth map from the real image(s), before comparing each pixel with the virtual depth to generate a compositing mask $C$. Bottom: We instead directly estimate the mask given the real image(s) and virtual depth as input. Additionally, our method also employs a lightweight temporal smoothing input to generate more stable predictions.
    }
    \label{fig:method_overview}
    \vspace{-7pt}
\end{figure*}

Our goal is to automatically composite virtual objects into images of real scenes, respecting any real occluding objects that are `in the way'.
At inference time, we assume we have an RGB image $I_\text{real}$ as input, together with a temporally preceding sequence of RGB source images and corresponding camera intrinsics and poses. 
We denote the full sequence of $I_\text{real}$ together with source images as $\mathcal{I}_\text{real}$.
We also assume knowledge of a 3D virtual object that we wish to place in the scene, which can change over time.
From the 3D virtual object and camera poses, we extract for each frame a color rendering of the virtual object $I_\text{virtual}$ with an associated virtual depth map $D_\text{virtual}$.

\subsection{Our approach}

Given the rendering of the asset, the job of an occlusion step is to estimate which virtual pixels should be shown, and which should be hidden, to create the final image $I_\text{final}$.
This can be described by a compositing equation \cite{porter1984compositing, smith1996blue}, using the two images and a per-pixel compositing mask $C$, so
\begin{align}
    I_\text{final} = C I_\text{real} + (1 - C) I_\text{virtual}.
    \label{eqn:compositing_method}
\end{align}
This compositing equation is only applied to pixels covered by the virtual object, outside of which we only show $I_\text{real}$.

Traditional occlusion methods, \eg \cite{ramamonjisoa2020predicting,valentin2018depth}, use a depth map to estimate $C$. The depth map $D_\text{real}$ is the output of a network $\psi$, which takes $\mathcal{I}_\text{real}$ as input, so $D_\text{real} = \psi({\mathcal{I}_\text{real}})$.
Here the compositing mask $C$ was formed using the relation
\begin{align}
    C = [D_\text{real} < D_\text{virtual}], %
\end{align}
where $[]$ is the Iverson bracket.

Training this network to instead \emph{directly} predict $C$ is a potentially attractive alternative solution. 
However, this direct prediction is not feasible without $D_\text{virtual}$, as the network has no context at inference time of where the virtual object should be positioned in the world and therefore would  be unable to produce a plausible mask.

In our approach, we instead use a deep network $\phi$ to directly estimate $C$, conditioned on \emph{both} $\mathcal{I}_\text{real}$ and $D_\text{virtual}$ as input, so 
\begin{align}
    C = \phi(\mathcal{I}_\text{real}, D_\text{virtual}).
\end{align}
The final image is then formed using Eqn.~\ref{eqn:compositing_method} from above.

\paragraph{Advantages of our depth informed mask prediction.}
We hypothesize that it is easier for our network to directly predict a \emph{binary} `in front vs.~behind' value at each pixel location, compared with existing methods that predict a \emph{continuous} value to regress the absolute depth.

\subsection{Predicting an occlusion map}
\label{sec:occlusion_map}
\newcommand{\loc}{\mathbf{p}}

A natural choice for our network $\phi$ would be an image-to-image network.
This would take the concatenation of $\mathcal{I}_\text{real}$ and $D_\text{virtual}$ as input, and then output $C$.
However, at training time such an architecture would need to see both realistic images and realistic \emph{virtual} depth maps corresponding to the scene.
Generating realistic virtual depths for a scene is difficult, as we do not know what the final use case of the system might be, and thus placing virtual objects in a scene automatically at training time is a non-trivial task.
We instead take a different approach, and propose an architecture with two parts: (i) a backbone network for image encoding, followed by (ii) a per-pixel MLP (see Figure~\ref{fig:method_overview}).
Our virtual depths are only provided to the per-pixel MLP, meaning our training-time virtual depths do not need to be realistic virtual depth maps.

\paragraph{Backbone network for image encoding.}
Our backbone network maps the RGB image $I_\text{real}$ to a pixel-aligned feature encoding $F$ with $K$ channels per pixel.
While we could use any backbone to extract features, for most of our experiments we use a multi-view stereo approach as in~\cite{sayed2022simplerecon}.
This requires temporally preceding source frames and known camera poses. %
See Section~\ref{sec:implementation_details} for details.

\paragraph{Predicting the occlusion mask with an MLP.}
The final prediction  of the compositing mask at pixel location $\loc$ relies on three inputs: 
\vspace{-6pt}
\begin{enumerate}
    \itemsep-3pt
    \item The image features at $\loc$, \ie $F(\loc)$. Inspired by \cite{kirillov2020pointrend}, we interpolate features from $F$ at arbitrary sub-pixel locations. This enables us to make final predictions at arbitrary locations and resolutions.
    \item The virtual object depth at $\loc$, \ie $D_{\text{virtual}} (\loc)$. Again, we can sample $D_\text{virtual}$ at arbitrary sub-pixel locations.
    \item The warped previous temporal prediction at $\loc$, as described next. 
    This enables the network to use temporal information for more  stable predictions.
\end{enumerate}
\vspace{-5pt}
At location $\loc$, we concatenate the above three inputs to make a $K+$2-dimensional feature.
This is given to an MLP to produce the final compositing output for that location $C (\loc)$.
The final layer of the MLP has a sigmoid activation, so $C (\loc)$ is continuous $ \in [0, 1]$. 
This can be interpreted as the probability that $I_\text{real}(\loc)$ is occluding the virtual object at depth $D_\text{virtual}(\loc)$.

\paragraph{Temporal stability.} 
Temporally stable occlusions are important for seamless and believable AR immersion as per-frame depth or semantic predictions can vary over time, leading to visual `flickering'.
To combat this, we encourage the network to be temporally stable, inspired by~\cite{tkachenka2019real, perazzi2017learning}.
To achieve this, we feed to the MLP the previous prediction for the pixel at location $\loc$, $\hat{C}(\loc)$, defined as $C^{t-1}(\textit{warp}[\loc])$. $\textit{warp}[]$ 
uses the known relative camera transform to backwards warp  the pixels' locations at time $t$ to time step $t-1$ using~\cite{jaderberg2015spatial}. This warping requires known depth at time $t$, for which we use the rendered \emph{virtual} depth. 
This is in contrast to \cite{tkachenka2019real} which does not use geometric information.

\paragraph{Relationship to prior work.}
While previous approaches (\eg \cite{badki2020bi3d}) have trained a network which takes a single query depth as input, our approach is novel as our query depth can vary spatially  per pixel, so each pixel can have a different query depth. 
Methods like \cite{badki2020bi3d} would require exhaustive evaluation of all depth values to be able to composite an object.
Our approach can be seen as producing `implicit depths', similar to prior work in 3D, \eg~\cite{park2019deepsdf}.

\subsection{Training our network}
\label{sec:training}

Our goal at training time is to update the weights of our network $\phi$ (\ie both feature encoder and MLP) to accurately predict occlusions.
We use training datasets, \eg \cite{dai2017scannet, hypersim}, where we have access to sequences of training images $\mathcal{I}_\text{real}$ with pixel-aligned ground truth depth maps $D_\text{real}$ and associated camera poses and intrinsics.
However, these datasets do not come with augmented virtual depths $D_\text{virtual}$, so we need to synthesize these at training time.

\paragraph{Our training samples.}
We require training tuples with an image location $\loc$, a virtual depth at that location $D_\text{virtual}(\loc)$, and a ground truth label $y_i \in \{0, 1\}$ stating if the virtual depth is in front (0) or behind (1) the real image depth map. Additionally, to encourage temporal stability, we require a previous prediction for this location, $\hat{C}(\loc)$.

To generate a training sample, we choose a single training sequence $\mathcal{I}_\text{real}$, with associated depth map $D_\text{real}$. 
We sample a 2D location $\loc$ uniformly in image space, and subsequently sample a training-time feature $F^{}(\loc)$.
Given the image location $\loc$, we have a choice to sample our synthesized virtual depth $D_\text{virtual} (\loc)$ anywhere along $\loc$'s camera ray.
Sampling a random depth means we might be far away from the difficult choices.
The most difficult choices for depths are when $D_\text{virtual}$ is near $D_\text{real}$, so we  bias a fraction of our training-time samples to come from near the ground-truth depth surface $D_\text{real}$. 
Similar to \cite{saito2019pifu,saito2020pifuhd}, with probability $q$ we sample from a Gaussian with mean of the ground truth depth value at pixel $\loc$, and variance 0.05. 
To ensure that we also make sensible predictions away from real surfaces, with probability $1-q$ we sample a training depth uniformly
\begin{wrapfigure}{r}{0.25\textwidth}
  \vspace{-6pt}
    \includegraphics[width=0.25\textwidth]{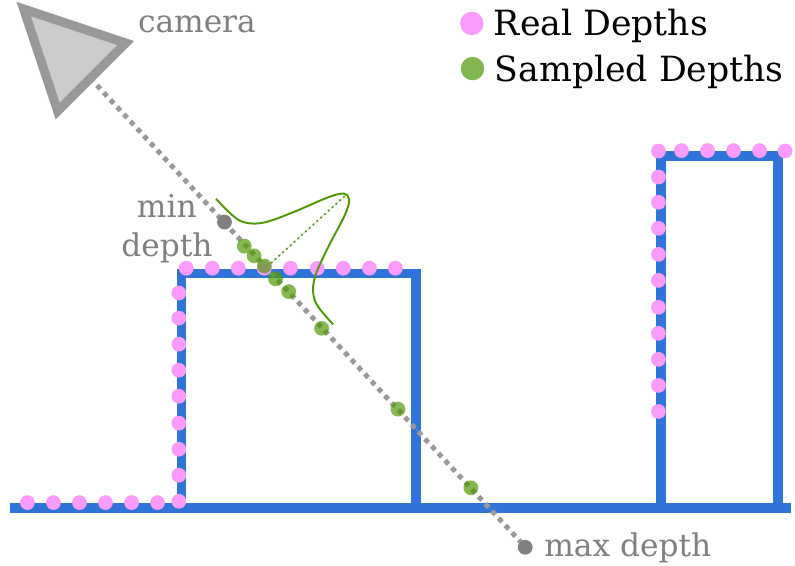}
   \vspace{-15pt}
\end{wrapfigure}
between the minimum and maximum depth in $D_\text{real}$.
The ground truth depth map determines the label $y_i$, which, in turn, is used to supervise the network with binary cross-entropy.

\begin{table*}
    \centering
    \footnotesize
    \resizebox{1.0\textwidth}{!}{
        \begin{tabular}{lcccccccc}
            \toprule
            
            \textbf{Method} & \multicolumn{3}{c}{\textbf{Occlusion evaluation}} &  \multicolumn{5}{c}{\textbf{Depth evaluation}} \\
            \cmidrule(lr){2-4}\cmidrule(lr){5-9} 
            
             & IoU All$\uparrow$ & IoU Surface$\uparrow$ & IoU Boundary$\uparrow$ & Abs Diff$\downarrow$ & Abs Rel$\downarrow$ & Sq Rel$\downarrow$ & RMSE $\downarrow$ & $\delta < 1.05\uparrow$ \\
            
            \midrule
            DPSNet ~\cite{im2019dpsnet} & 46.17 & 21.68 & 23.50 & .1552 & .0795 & .0299 & .2307 & 49.36  \\
            MVDepthNet ~\cite{wang2018mvdepthnet} & 44.64 & 21.06 & 23.22  & .1648 & .0848 & .0343 & .2446 & 46.71 \\
            DELTAS~\cite{sinha2020deltas} & 48.48 & 23.37 & 25.51  & .1497 & .0786 & .0276 & .2210 & 48.64\\
            GPMVS ~\cite{hou2019multi} & 46.52 & 22.43 & 23.98  & .1494 & .0757 & .0292 & .2287 & 51.04 \\
            DeepVideoMVS, fusion~\cite{duzceker2021deepvideomvs}*  & 53.16 & 26.49 & 28.05 & .1186 &  .0583 &  .0190 & .1879  &  60.20 \\
            SimpleRecon (ResNet) & 58.91 & 31.48 & 33.03 & .0978 & .0487 & .0151 & .1617 & 69.62 \\
            SimpleRecon \cite{sayed2022simplerecon} & \underline{60.52} & 32.44 & 34.52 & \underline{.0871} & \textbf{.0429} & \underline{.0125} & \underline{.1460} & \textbf{74.01}\\
            
            \midrule
            
            SimpleRecon (ResNet) + \textbf{Ours} & 60.14 & \underline{33.29} & \underline{36.54} &  .0988 &  .0498 &  .0149 & .1595 &   68.52 \\
            SimpleRecon \cite{sayed2022simplerecon} + \textbf{Ours}  & \textbf{62.61} & \textbf{35.48} & \textbf{38.01}  & \textbf{.0862} & \underline{.0436} & \textbf{.0123} & \textbf{.1426} &\underline{73.74}
            \\

            \bottomrule
            
        \end{tabular}
    }
    
    \vspace{1pt}
    \caption{\textbf{Occlusion and depth scores after converting our masks to depths compared with state-of-the-art prior works.}  Evaluation is on the ScanNetv2 test set keyframes~\cite{duzceker2021deepvideomvs}, and follows the evaluation protocol for \emph{depth} from~\cite{duzceker2021deepvideomvs}. Our model is state-of-the-art on both occlusion and depth estimation. * indicates trained on additional data.}
    \vspace{-4pt}
    \label{tab:depth-evaluation}
\end{table*}

\begin{figure}
    \centering
    \includegraphics[width=1.0\columnwidth]{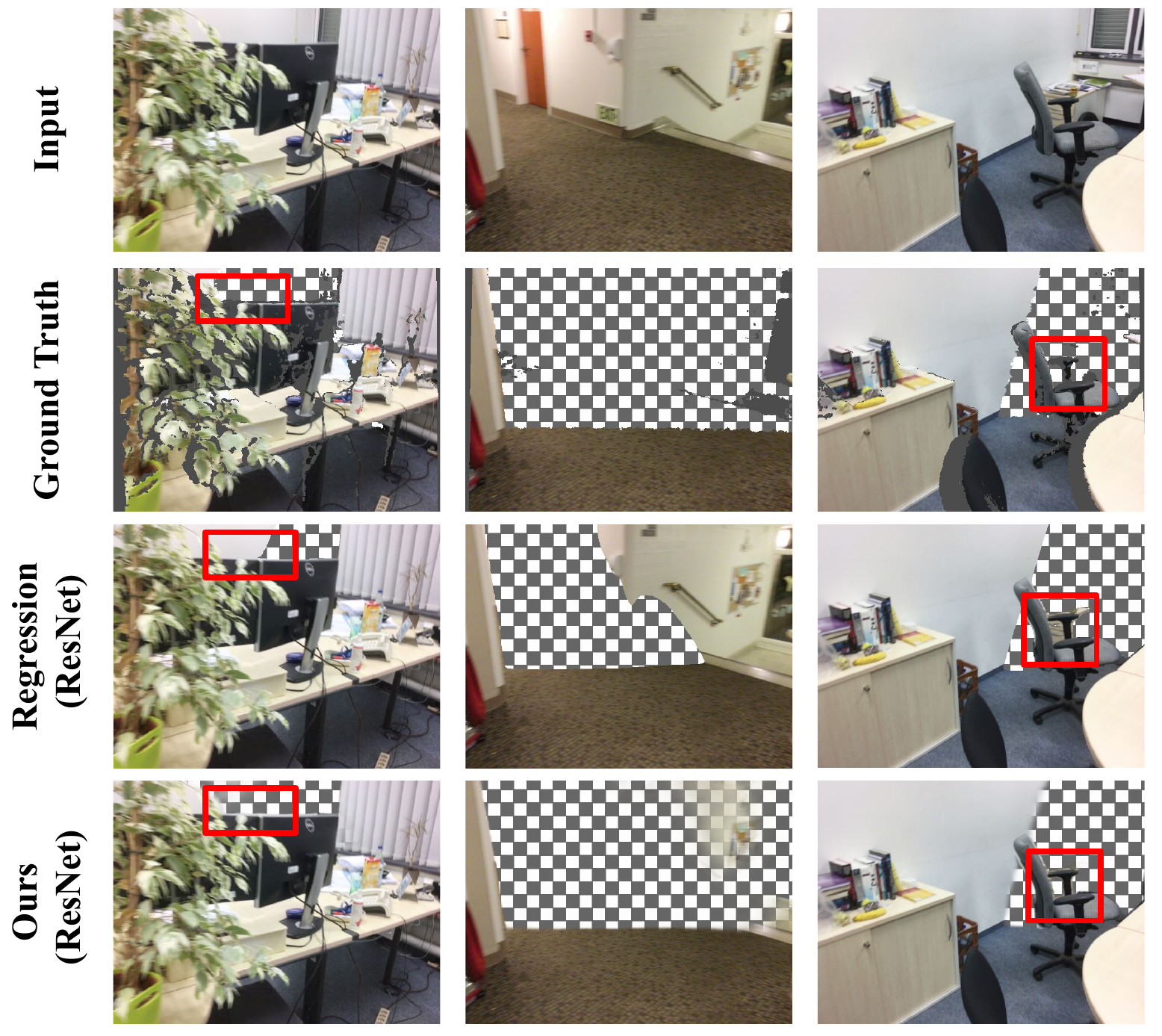}
    \caption{\textbf{Occlusion predictions on ScanNetv2.} Comparison of occlusion masks when a virtual plane sits at 3m from the camera. ScanNetv2's ground truth is noisy, especially near thin structures and distant points $>$ 5m. Regression is crisp but makes mistakes affecting whole regions. Ours misclassifies part of a painting in the example in the second column, but is better overall. %
    }
    \label{fig:planes_qual}
    \vspace{-10pt}
\end{figure}

\paragraph{Training for temporal stability.}
As stated in Section~\ref{sec:occlusion_map}, we encourage temporal stability by giving the warped previous prediction as an additional input to our MLP. During training, to avoid the need to run inference for multiple frames, we synthesize a previous prediction from the ground truth label $y_i$ in a manner similar to \cite{tkachenka2019real}. We corrupt $y_i$ to produce a pseudo previous prediction $\Tilde{C} (\loc)$ by adding random noise, converting the binary labels into floats $\in [0, 1]$, simulating the output of a sigmoid.  
To teach our model to be robust to incorrect previous predictions at inference time, we set $\Tilde{C} (\loc) = 1 -  \Tilde{C} (\loc)$ with probability $p_1$ during training. Additionally we use $\Tilde{C} (\loc) = -1$ to indicate the start of a sequence, and set this during training with probability $p_2$. 
For all experiments we use $p_1 = p_2 = 0.25$.

\paragraph{Regularization around depth discontinuities.}
Training a model as described in Section~\ref{sec:training} can give rise to artefacts in predicted compositing masks near depth discontinuities. 
Due to the inherent ambiguity of occlusions in these regions, our model tends to predict values close to 0.5, which leads to less visually pleasing results in the final compositing.
To combat this, during training we locate depth discontinuities in the ground truth depth map using a Sobel filter~\cite{kanopoulos1988design}, and apply an L1 regularizer to penalize predictions near 0.5 in these regions.
See the supplementary material for details.

\begin{table}
    \centering
    \resizebox{.48\textwidth}{!}{%
    \begin{tabular}{lccc}
        \toprule
        \textbf{Architecture / Method} & \textbf{IoU All$\uparrow$}  & \textbf{IoU Surface$\uparrow$}  & \textbf{IoU Boundary$\uparrow$}   \\
        
        \midrule
        
        SimpleRecon \cite{sayed2022simplerecon} & 
        {60.52} & {32.44} & {34.52} 
        \\
        
        SimpleRecon \cite{sayed2022simplerecon} + \textbf{Ours} & \textbf{62.61} & \textbf{35.48} & \textbf{38.01} 
        \\
        
        \midrule
        
        SimpleRecon (ResNet)  & 
        {58.91} & {31.48} & {33.03} 
        \\
        
        SimpleRecon (ResNet) + \textbf{Ours} & 
        \textbf{60.14} & \textbf{33.29} & \textbf{36.54}
        \\
        
        \midrule
        
        ManyDepth \cite{watson2021multidepth} & 
        {55.68} & {29.18} & {30.47} 
        \\
        
        ManyDepth \cite{watson2021multidepth} + \textbf{Ours} & 
        \textbf{56.55} & \textbf{31.56} & \textbf{34.47}
        \\
        
        \midrule
        
        MonoDepth2 \cite{godard2019digging} & 
        {46.06} & {18.62} & {23.37} 
        \\
        
        MonoDepth2 \cite{godard2019digging} + \textbf{Ours} & 
        \textbf{48.35} & \textbf{21.55} & \textbf{26.55} 
        \\

        \bottomrule
    \end{tabular}
    }
    
    \vspace{2pt}
    \caption{\textbf{Our method outperforms depth regression for the occlusion task,} regardless of the underlying architecture.
    All these architectures were trained and evaluated on ScanNetv2 sequences by us, using code published by the authors, and with automated virtual object insertion to evaluate binary occlusion masks. 
    }
    \vspace{-4pt}
    \label{tab:virtual-asset-evaluation}
\end{table}

\subsection{Implementation details}
\label{sec:implementation_details}
We train all models and baselines using the Adam optimizer \cite{adamsolver} with a batch size of 24 split across 2 GPUs. For speed of convergence, we initialize our backbone network with the weights of a depth regression network, and train for 40k steps, with an initial learning rate of 0.0001 dropping by a factor of 10 after 16k steps and 32k steps.
Images are augmented with standard flip and color augmentations, as in~\cite{sayed2022simplerecon}. We use $q = 0.25$ for our probability of sampling a virtual depth near the real depth surface.
Similar to~\cite{godard2019digging}, we supervise the network at 4 output scales, using ground truth depth at a higher resolution than the feature maps, and at test time only use the highest resolution output. 

\paragraph{Backbone.}
Our backbone network is based on~\cite{sayed2022simplerecon}, where a shallow feature extractor feeds a metadata infused plane sweep volume, followed by a U-Net \cite{ronneberger2015u}. 
See the supplementary material for full details.

\paragraph{MLP.}
Our compositing mask prediction uses an MLP with three fully-connected layers, with $K+$2 input channels and a single channel final output. Both hidden layers have 128 dimensions.
We use ELU activations after the first two layers, with the final activation being a sigmoid to map our outputs to the range $[0, 1]$. 
Based on our backbone \cite{sayed2022simplerecon}, our full-scale feature map has $K=64$.

\paragraph{Timings.} Inference with our backbone (from~\cite{sayed2022simplerecon}) takes 64ms and our MLP takes 0.5ms on an A100 GPU.

\section{Experiments}
\label{sec:experiments}

\paragraph{Datasets.}
We train and evaluate models using ScanNetv2~\cite{dai2017scannet}, with the standard train/val/test split.
This allows direct comparison, of occlusions or depths, with most prior depth estimation methods that also train on ScanNetv2.
For qualitative comparisons, we also train a model on the synthetic Hypersim~\cite{hypersim} dataset.
This synthetic data has better aligned edges in the training data, so yields a model with high edge fidelity.
In visual occlusion comparisons, we compare that model against Lidar depth sensing~\cite{arkit} and a previous method which does not use a trained model~\cite{holynski2018fast}.

\paragraph{Backbone variants.} 
We present two variants of our model; `SimpleRecon + \textbf{Ours}' which uses the architecture of~\cite{sayed2022simplerecon} as a backbone, and a faster lower compute version using a ResNet-18~\cite{he2016deep} encoder, a lightweight decoder, and a simple dot-product cost volume, referred to as `SimpleRecon (ResNet) + \textbf{Ours}'. We also train a ResNet variant \emph{without} a cost volume inspired by~\cite{godard2019digging}: `MonoDepth2 + \textbf{Ours}'.

\subsection{Evaluating virtual asset occlusion}
\label{sec:occlusion_evaluation}

We directly evaluate performance on the task of virtual object insertion. We report scores using the standard segmentation metric, intersection-over-union (IoU), to measure occlusion quality. We compare our variants against state-of-the-art depth estimation on the standard ScanNetv2 test set.  

Since rendered virtual assets would add noise to the evaluation process, we use infinite planes that lie ahead of the camera at each frame. 
These planes are placed at depths $d \in \{0.5\text{m}, 1.0\text{m}, \cdots, 5.0\text{m}\}$ along the look-at vector of the camera, where each plane's virtual depth map is $D_\text{virtual}^d$. 
We obtain a ground truth binary occlusion mask, $Y_\text{GT}^d$, for each plane using the ground truth depth map, obtained from depth sensors.

For our method, we compute the probability of occlusion for each depth plane, $C^d$, which we threshold with $\tau$ to produce $Y^d_\text{pred}$. 
We pick $\tau$ for each depth bin using a mini-val set of 100 scans. 
We compute IoU for the occluded asset fragments, $\text{IoU}_-^{d}$, and the visible parts of the asset, $\text{IoU}_+^{d}$. 
For depth estimation methods, we obtain the predicted occlusion mask directly by comparing $D_\text{pred}^d$ and $D_\text{virtual}^d$ to compute $\text{IoU}_+^{d}$ and $\text{IoU}_-^{d}$.
We compute $\text{IoU All}^{d}$ for each plane using the harmonic mean of $\text{IoU}_+^{d}$ and $\text{IoU}_-^{d}$.
We average IoUs for each keyframe from \cite{duzceker2021deepvideomvs} and then for each depth plane.
As regions near depth boundaries tend to be difficult, following~\cite{chen2017deeplab} we evaluate \emph{IoU Boundary}, and separately, regions near the geometry's surface (\emph{IoU Surface}).

In Table~\ref{tab:virtual-asset-evaluation}, we combine our method with existing backbones by first training with regression losses \cite{sayed2022simplerecon}, and then finetuning with our approach on ScanNetv2. We compare with several recent MVS methods, including SimpleRecon \cite{sayed2022simplerecon}, ManyDepth \cite{watson2021multidepth}, as well as a single frame depth method MonoDepth2~\cite{godard2019digging}. In all cases, our method improves occlusion scores, most notably in difficult cases near surfaces. Additionally, our lightweight ResNet variant yields strong performance (sometimes exceeding \cite{sayed2022simplerecon}) while operating at a fraction of the compute time (20ms vs 64ms on an A100 GPU). 
Note that for fair comparison to regression baselines, all results (except those in Table~\ref{tab:temporal}) are without our temporal stability contribution.

\begin{figure}
    \centering
    \includegraphics[width=1.0\columnwidth]{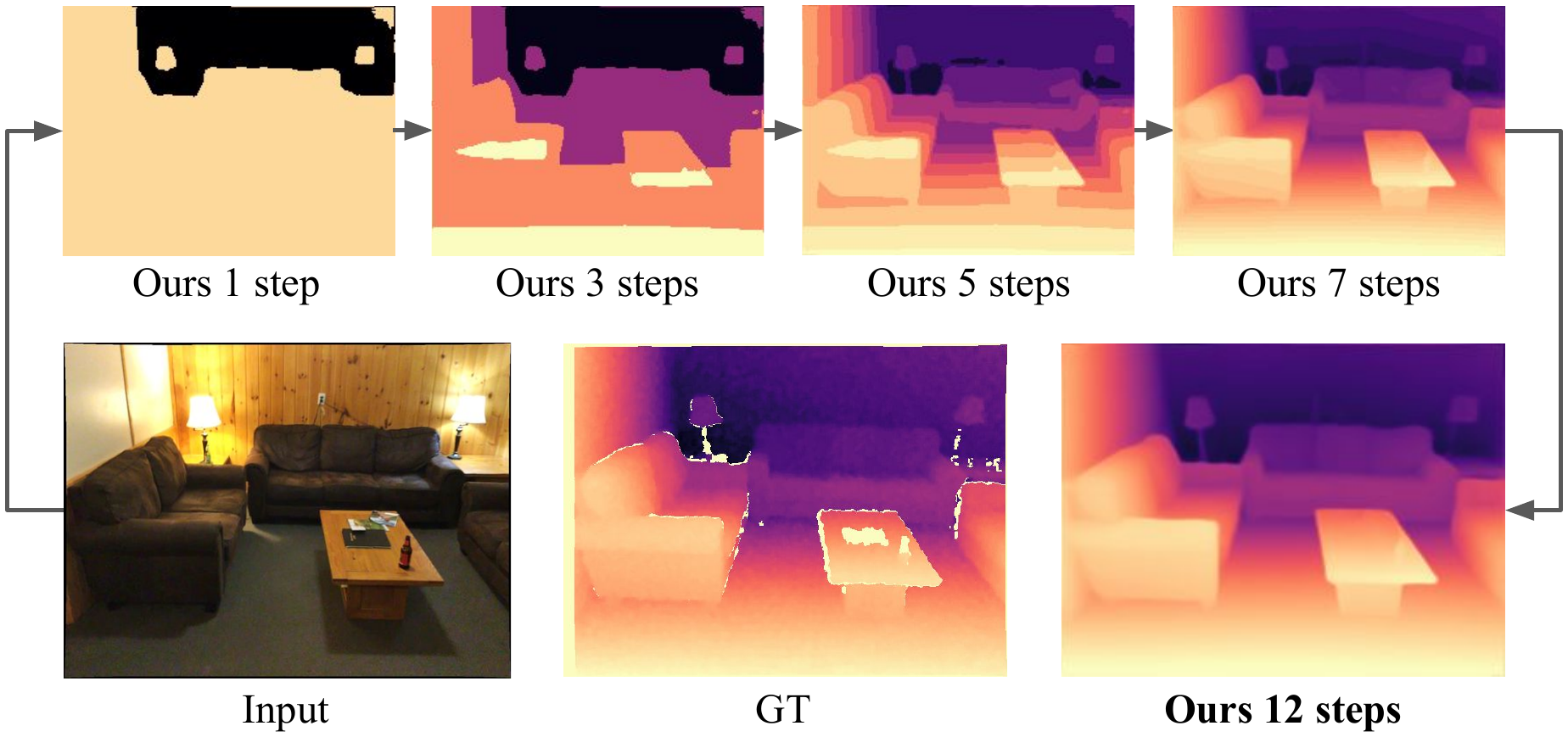}
    \caption{\textbf{Our binary search converts our predictions to a full depth map.} First the backbone is run only once to extract features, then the much faster binary prediction MLP is run once for each step of binary search, iteratively refining a depth map.%
    }    
    \label{fig:binary_search}
    \vspace{-5pt}
\end{figure}

\begin{table*}
    \centering
    \footnotesize
    \begin{tabular}{lccccccc}
        \toprule

        \textbf{Method} & \multicolumn{3}{c}{\textbf{Occlusion evaluation}} &  \multicolumn{4}{c}{\textbf{Depth evaluation}} \\
        \cmidrule(lr){2-4}\cmidrule(lr){5-8} 
        
         & IoU All$\uparrow$ & IoU Surface$\uparrow$ & IoU Boundary$\uparrow$ & Abs Rel$\downarrow$ & Sq Rel$\downarrow$ & RMSE $\downarrow$ & $\delta < 1.05\uparrow$ \\

        \midrule

        Ours (no edge-regularization)  & 59.79 & 33.03 & 36.02 & 
          .0503 & .0155 & .1617 & 68.31
        \\
        
        Ours (no high resolution supervision)  & 59.94 & 33.09 & 36.31 & .0501           & .0157 & .1628 & 68.44        \\

        Ours (monocular backbone)   & {48.24} & {21.46} & {26.72} & .1220 & .0533	& .2740 & 34.71
        \\
        
        DORN-style classification  & {56.81} & {29.54} & {31.29} &  .0535 & .0171 & .1743 & 65.41
        \\
        
        Classification   & {55.95} & {29.49} & {31.44} &  .0617 & .0220	& .1937 & 60.75
        \\
        Regression & 58.91 & 31.48 & 33.03 & \textbf{.0487} & .0151 & .1617 & \textbf{69.62} \\

        \midrule
        
        Ours & \textbf{60.14} & \textbf{33.29} & \textbf{36.54} & 
        .0498 & \textbf{.0149} & \textbf{.1595} & 68.52
        \\
        
        \bottomrule
        
    \end{tabular}
    
    \vspace{1pt}
    \caption{\textbf{Ablating our method, showing our contributions lead to better depth and occlusion scores.} All are trained equivalently on the ScanNetv2 dataset using the SimpleRecon (ResNet) backbone network.}
    \vspace{-5pt}
    \label{tab:ablation}
\end{table*}

\subsection{Evaluating depth estimation}
While our method is focused on estimating occlusions for virtual assets, we can leverage our binary predictions to iteratively refine a binary-searched depth map. We compare against depth estimation methods on the ScanNetv2 test set using the protocol from \cite{duzceker2021deepvideomvs}, presenting results in Table~\ref{tab:depth-evaluation}. 

We can convert our binary predictions to depths by making the observation that along each ray from every pixel location $\loc$ there lies a depth $d$ where the prediction from our network, $C(\loc)^d$, is at the decision threshold $\tau$. Generally $\tau$ would be 0.5, but we use the best thresholds from Section~\ref{sec:occlusion_evaluation}. 
Our optimal estimated depth map $D_\text{pred}$ is one where $C(\loc)^d = \tau$ for all $\loc$.

It is time consuming to naively iterate different depth values to find $d$ for which $C(\loc)^d = \tau$. Instead we binary search along the ray to find the optimal depth, relying on directional predictions that signify if the current depth on the ray is ahead or behind the real depth. We initialize our min.~and max.~depths to 0.5m and 8m respectively. 
These are updated each iteration, for each location $\loc$, as we search. 
We take $M=12$ binary search steps, achieving an effective granularity of 4096 for each $\loc$ (see Figure~\ref{fig:binary_search}). 
Only the final MLP head is run at each step, with the backbone only run once to produce feature maps that are reused.

Notably, when using \cite{sayed2022simplerecon} as a backbone, our method achieves a new state-of-the-art on ScanNetv2 in depth estimation, alongside our core occlusion-IoU evaluation.

\subsection{Temporal evaluation}
Occlusion systems should exhibit \emph{temporal coherence} to ensure visually compelling results~\cite{holynski2018fast}.
In the spirit of~\cite{luo2020consistent, holynski2018fast}, we place a fixed virtual AR asset into a scene (here a plane), and keep track of the change in predicted visibility of the groundtruth scene mesh, provided in ScanNetv2, across a window of frames.
Specifically, we use an infinite vertical plane at a fixed position in front of the first camera in a sequence, and compute a compositing mask, $C^t$, for each subsequent frame. 
We project scene mesh vertices to the camera and store the visibility prediction from $C^t$ for that vertex w.r.t the vertical plane. We tally the number of times the visibility prediction changes for each vertex over 13 frames (\ie~0.43 seconds). 
We normalize the count by the number of frames to compute a temporal score. 
Evaluation is performed on the ScanNetv2 test scenes. 

\begin{figure}
    \centering
    \includegraphics[width=1.0\columnwidth]{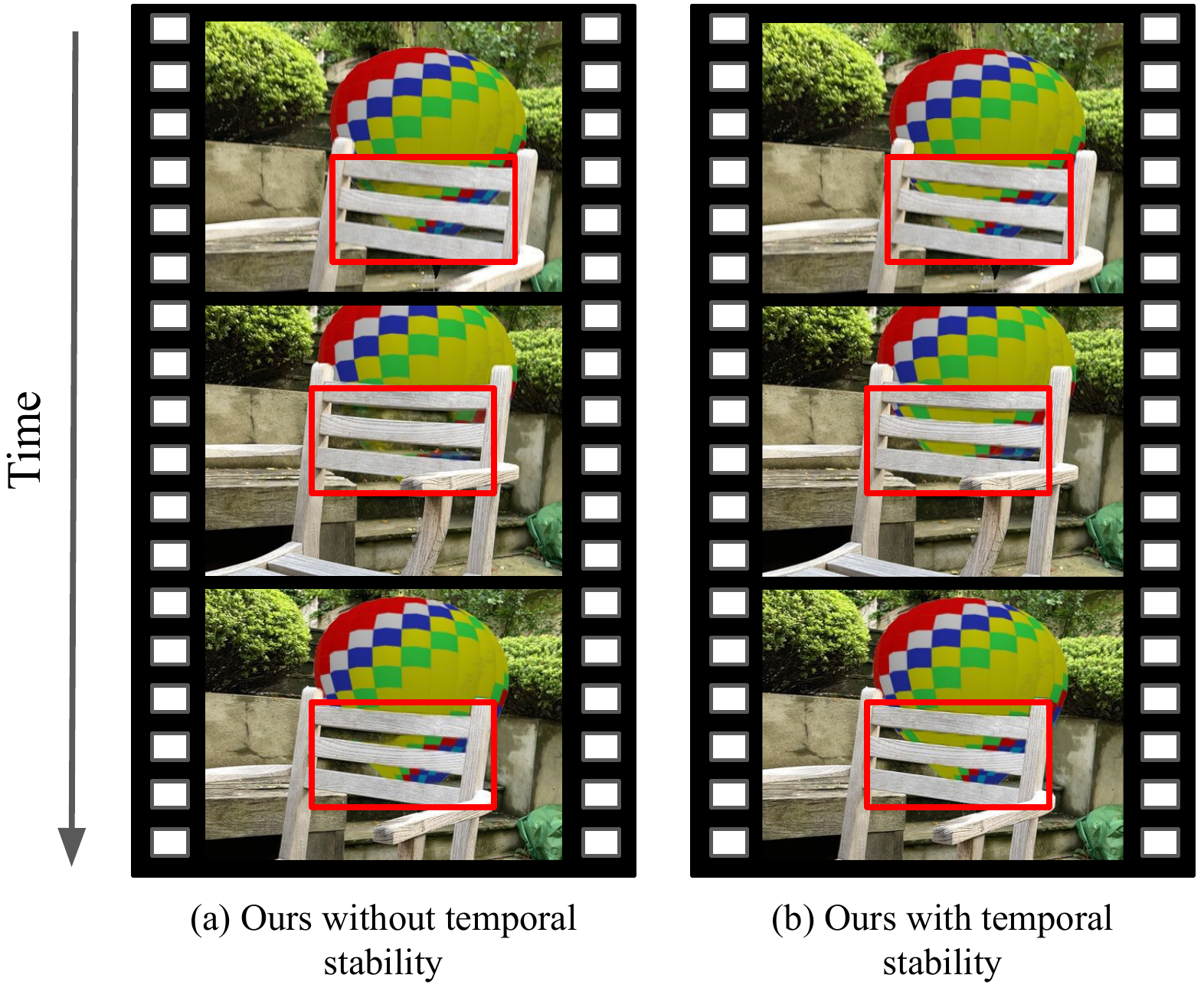}
    \caption{\textbf{Temporal stability.} (Left) The basic method without our temporal stability input displays prominent flickering  \ie big changes in the predictions for each frame of this sequence. (Right) Our predictions are more temporally stable over time, enabling more immersive AR. Please also see the accompanying video. }
    \label{fig:temporal_qualitative}
    \vspace{-10pt}
\end{figure}

`Ours (with temporal)' results in  a significant boost of almost 30\% in temporal stability while achieving IoU scores comparable to our non-temporal variant (see Table~\ref{tab:temporal}).
We also show a qualitative example of our more temporally stable approach in Figure~\ref{fig:temporal_qualitative}. Please see our video for examples.

\begin{table}
    \centering
    \footnotesize

    \resizebox{1.0\columnwidth}{!}{
        \begin{tabular}{lcccc}
            \toprule
            \textbf{Method} & \textbf{Temporal Score$\downarrow$} & \textbf{IoU A.$\uparrow$} & \textbf{IoU S.$\uparrow$} & \textbf{IoU B.$\uparrow$} \\
            \midrule

            Regression  & 233.1 & 77.84 & 43.44 & 41.29 \\
            Ours (w/o temporal)  & 235.1 & 79.09 & \textbf{44.73} & 42.44 \\
            \midrule
            Ours (with temporal)  & \textbf{164.5} & \textbf{79.28} & 44.50 & \textbf{42.90} \\
            
            \bottomrule
        \end{tabular}
    }
    \vspace{1pt}
    \caption{\textbf{Evaluating temporal stability on ScanNetv2}, by comparing predictions on temporally adjacent frames. Our temporal approach leads to significantly less flicker, as seen in the large reduction in the temporal score, without impacting IoU.}
    \label{tab:temporal}
    \vspace{-10pt}
\end{table}

\subsection{Ablation}
We validate our approach by training variants of our model with our contributions turned off, and show in Table~\ref{tab:ablation} that these models achieve worse scores.
We train a model without our edge-based regularization; without high resolution supervision, where our MLP is supervised at the native model output resolution as in \cite{sayed2022simplerecon}; a non-MVS monocular method;  a DORN-style \cite{fu2018deep} classification network, where we output a classification head with 80 bins each with a BCE-trained sigmoid activation; and a discretized depth-classification loss.

\begin{figure*}
    \centering
    \footnotesize
    \newcommand{\depthwidth}{0.2\textwidth}
    \setlength{\tabcolsep}{1pt}
    \begin{tabular}{cccccc}
        \textbf{Input} & 
        \textbf{ARKit Lidar} & 
        \textbf{Depth Dens. \cite{holynski2018fast} w/ Lidar } & 
        \textbf{SimpleRecon \cite{sayed2022simplerecon}} & 
        \textbf{\textbf{SR + Ours} } \\
        
        \input{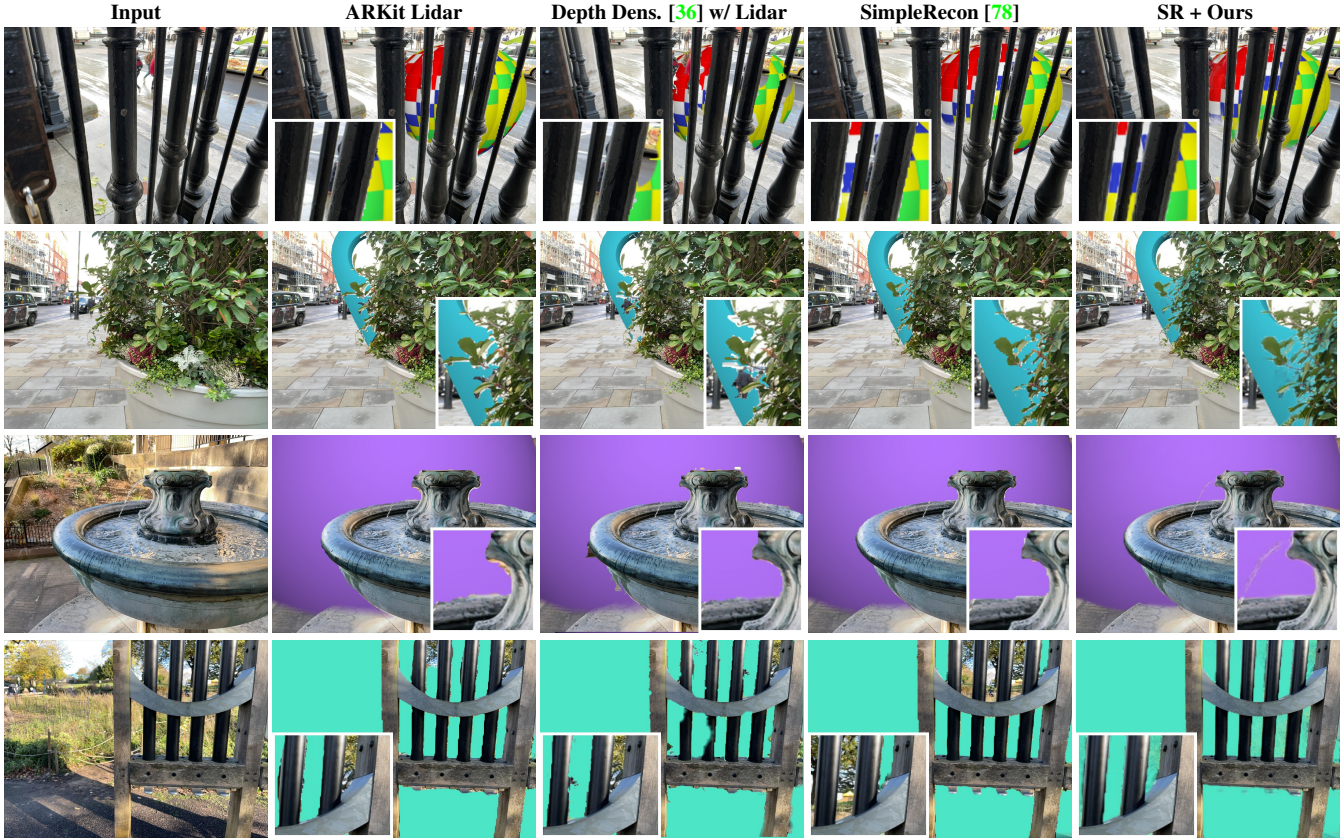}

    \end{tabular}
    \caption{\textbf{Qualitative occlusion comparisons using our own casually captured footage.} We occlude virtual assets (rows 1-2) and a fixed plane at 2m depth (rows 3-4). Our occlusions are typically more realistic than baselines, in particular around soft edges, \eg leaves. We also avoid catastrophic failures, \eg around the bars in the final row. Please see the supplementary material for videos.}
    \label{fig:qualitative_occlusions}
    \vspace{-10pt}
\end{figure*}

\begin{figure}[ht]
    \centering
    \footnotesize
    \includegraphics[width=1.0\columnwidth]{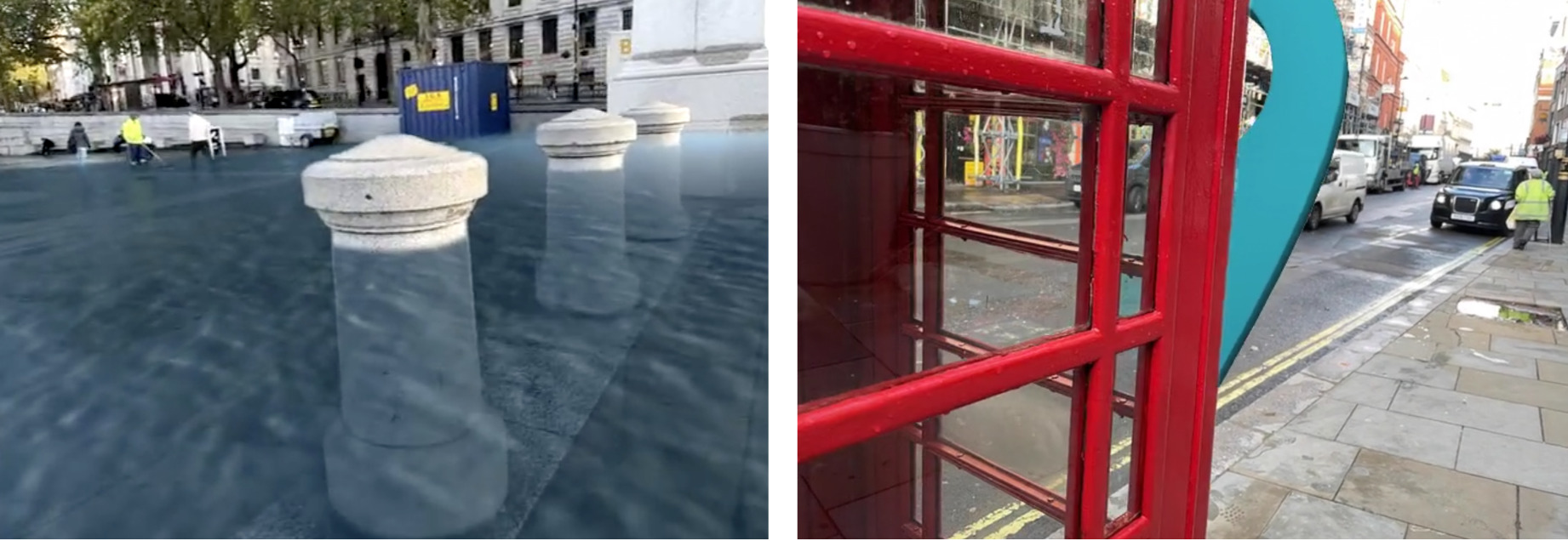}

    \caption{\textbf{Additional qualitative results.} On the right we see a failure mode, where transparency through glass is not handled correctly. This is due to limitations in our training data~\cite{hypersim}. Please see our video for more results.
    \label{fig:video_teaser}}
    \vspace{-3pt}
\end{figure}

\begin{figure}
    \centering
    \footnotesize
    \includegraphics[width=\linewidth]{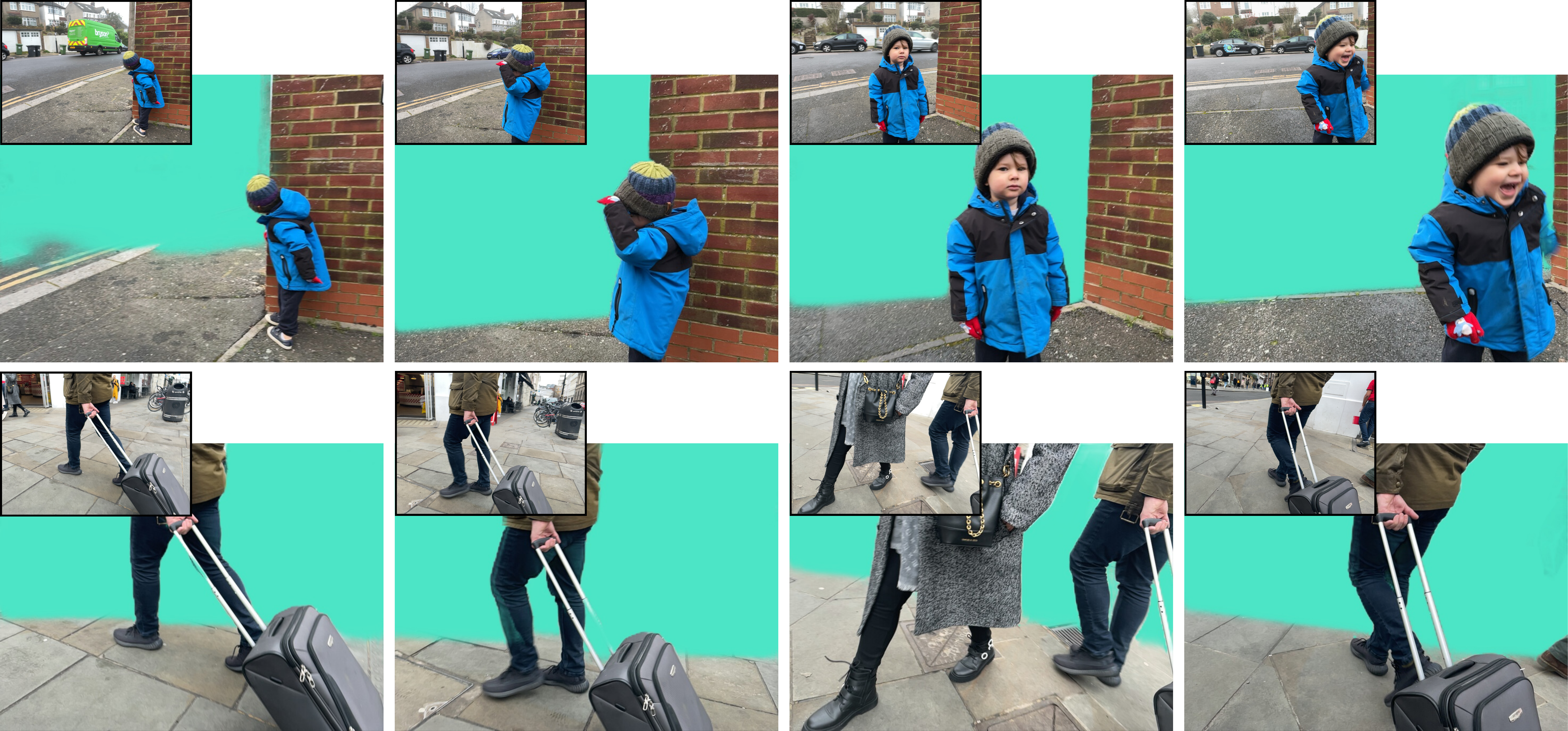}
    \caption{\textbf{Moving objects.} While our model was trained on static scenes, we achieve surprisingly robust results on moving objects. 
    \label{fig:moving_objects}}
    \vspace{-13pt}
\end{figure}

\subsection{Qualitative comparisons}
Figure~\ref{fig:qualitative_occlusions} shows qualitative results trained on Hypersim~\cite{hypersim} on a range of real-world scenes, comparing our approach to alternative state-of-the-art methods.
Surprisingly, our method produces visually equivalent, or better-quality, predictions compared to those from on-device Lidar from an iPhone 12 Pro.
We also compare to the sparse-point densification approach from \cite{holynski2018fast}.
Their approach relies on sparse points as input.
We found that ARKit's SLAM points are too sparse for their method, so we instead randomly sampled 2,000 Lidar depth points for each test frame, and fed these into their publicly available code.
Visually, our results have better edge fidelity than their Lidar-guided predictions.
We also found that our predictions are surprisingly good on moving objects, given that our training data comes from static scenes (see Figure~\ref{fig:moving_objects}).
More results are shown in Figures~\ref{fig:planes_qual} and \ref{fig:video_teaser}, and the supplementary video.

\section{Conclusion}
We presented a novel approach for inserting virtual objects into real scenes. 
In contrast to existing depth-based methods, we directly estimate compositing masks.
We introduce metrics for evaluating occlusion mask quality, and showed that our approach allows for greater temporal stability than previous methods. 
Qualitative results highlight that our method produces more realistic object insertions. 
A natural continuation of our work is to train a model in a fully `end-to-end' fashion, where the network directly outputs the final composited image $I_\text{final}$.
This could allow the network to reason about lighting and shadows \cite{legendre2019deeplight, tarko2019real} and object positioning~\cite{breen1996interactive}.

\vspace{2pt}
{
\noindent{\bf Acknowledgements.} Many thanks to Daniyar Turmukhambetov, Jamie Wynn, Cl\'{e}ment Godard, and Filippo Aleotti for their valuable help and suggestions.
}

\clearpage

{\small
\bibliographystyle{ieee_fullname}
\bibliography{main}
}

\clearpage
\appendix
\setcounter{table}{0}
\renewcommand{\thetable}{A\arabic{table}}
\setcounter{figure}{0}
\renewcommand{\thefigure}{A\arabic{figure}}
\noindent{\LARGE Supplementary Material}

{
  \hypersetup{linkcolor=blue}
  \tableofcontents
}

\section{Comparing ScanNetv2 vs.~Hypersim models}

\begin{figure*}[!hbt]
    \centering
    \newcommand{\depthwidth}{0.2\textwidth}
    \footnotesize
    \setlength{\tabcolsep}{1pt}
    \begin{tabular}{cccccc}
        \textbf{Input} &
        \textbf{Regression (Hypersim-trained)} &
        \textbf{Ours (Hypersim-trained)} &
        \textbf{Regression (ScanNetv2-trained)} & 
        \textbf{Ours (ScanNetv2-trained)} \\
        \input{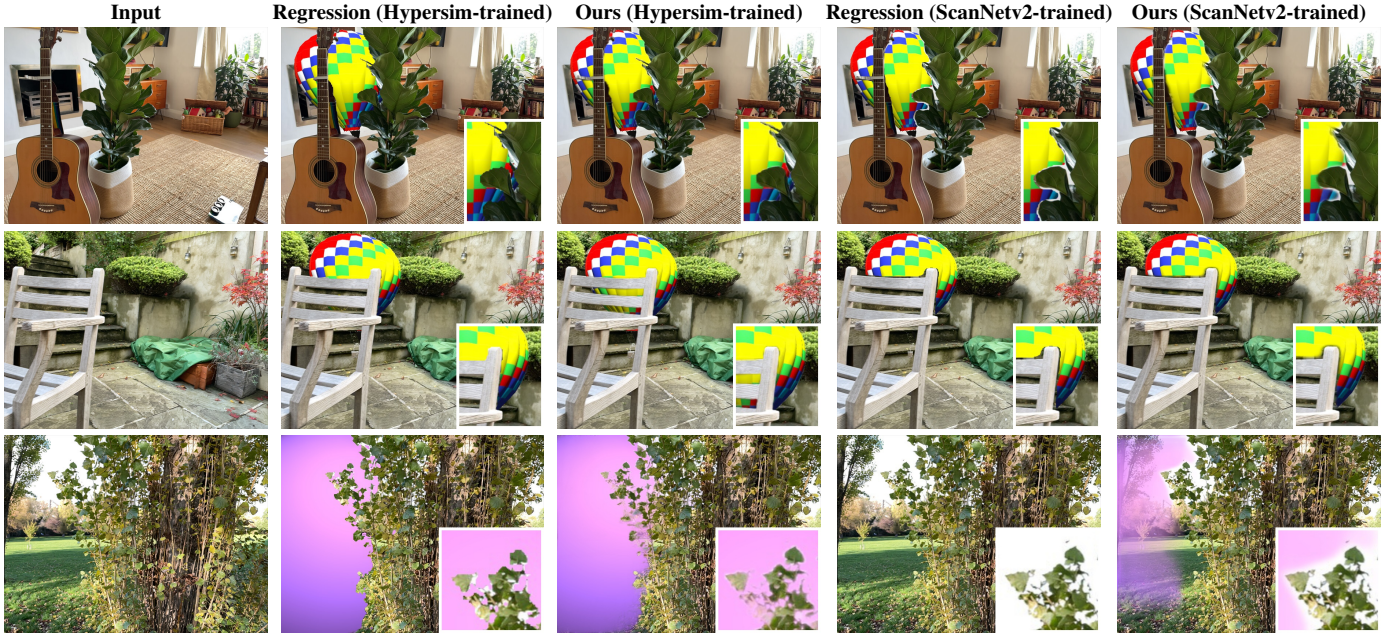}
    \end{tabular}
    \caption{
        \textbf{Comparing models trained on Hypersim versus ScanNetv2.}  Hypersim results in superior edges and better generalization on real data. Models trained on ScanNetv2 tend to have wider edges around objects and more catastrophic failures on outdoor scenes. 
        Regardless of the training data used, \textbf{models with our contributions outperform regression models}, as can be seen here.
        \label{fig:hypersim_vs_scannet}
        }
\end{figure*}

In Figure~\ref{fig:hypersim_vs_scannet}, we compare models trained on ScanNetv2 with models trained on Hypersim.
Compared with the ScanNetv2 model, we observe that Hypersim-trained models generally have better edge fidelity, and better generalization to outdoor scenes.
However, irrespective of the choice of training data, \emph{models trained with our contributions outperform the depth regression baselines}.

\section{Additional experiments}

\subsection{Additional temporal stability experiments}

In Table~\ref{tab:temporal_sup_mat} we present additional temporal stability evaluation.
Specifically, we compare:  %
\begin{itemize}
    \itemsep-2pt
    \item Our regression baseline, evaluated on our temporal stability metrics. 
    \item A version of our regression baseline, \emph{with} temporal stability added. In spite of several attempts with different settings, we found that adding versions of temporal stability to regression consistently made temporal scores worse.    
    In the table we show the best-performing version, where we add one extra input channel to the final decoder block at each scale. We use this additional input channel to feed the previous prediction to try to encourage temporal stability. At training time, we use the ground truth depth to create a pseudo previous prediction as in our implicit depth segmentation approach. At test time, we feed the network the previous prediction warped to the current frame as we do for our implicit depth model. We additionally show results for the same model but with no previous predictions at test time (indicated by setting the previous prediction input to $-1$).  %
    \item DVMVS~\cite{duzceker2021deepvideomvs}, a regression baseline that uses an LSTM. This should be considered a good option for temporally stable predictions. However, we found that it does not score well as it tends to flicker over time.
    \item Our main model, with temporal stability at training and test time (by feeding the MLP the previous warped prediction). 
    \item Our main model, without temporal stability. 
    \item Our main model, with temporal stability at training time but not during evaluation.  At test time we feed $-1$ as the previous prediction input to the MLP for every test image. We typically use $-1$ as a signal to the MLP at training time to indicate that the  previous prediction is unavailable. In this setting, the model never gets to see the previous frame. 
    \item A variant of our model, with a re-implementation of the augmentation method from \cite{tkachenka2019real}, applied at both training and test time.
    At training time, we simulate `previous' predictions by feeding into the MLP a copy of the current ground truth output.
    At test time, we simply feed the previous (un-warped) prediction.
    This is effectively our method but without our contributions of (a) warping the previous prediction using the camera motion, and (b) the addition of noise to simulate the output of a sigmoid.
    
\end{itemize}

\begin{table*}
    \centering
    \footnotesize

        \begin{tabular}{lcccc}
            \toprule
            Method & Temporal Score$\downarrow$ & IoU All$\uparrow$ & IoU Surface$\uparrow$ & IoU Boundary$\uparrow$ \\
            \midrule
            
            Regression (w/o temporal)  & 233.1 & 77.84 & 43.44 & 41.29 \\
            Regression (temporal)  &  262.1 & 75.75 & 42.04 & 39.90 \\
            Regression (temporal, with $-1$ at test time)  &  262.2 & 75.72 & 41.96 & 39.83 \\
           DVMVS~\cite{duzceker2021deepvideomvs} & 249.6 & 68.35 & 34.55 & 32.87 \\
            \midrule 
            Ours (temporal)  & \textbf{164.5} & \textbf{79.28} & 44.50 & \textbf{42.90} \\
            Ours (w/o temporal)  & 235.1 & 79.09 & 44.73 & 42.44 \\
            Ours (temporal, with $-1$ at test time) & 259.4 & 78.91 & 44.12 & 42.30 \\
            Ours (temporal from \cite{tkachenka2019real}) & 225.9 & 78.65 & \textbf{45.04} & 42.16 \\
            
            \bottomrule
        \end{tabular}
    \vspace{3pt}
    \caption{\textbf{Additional evaluation of temporal stability on ScanNetv2}. This table is comparable to Table 3 in the main paper.}
    \label{tab:temporal_sup_mat}
\end{table*}

\subsection{Additional ablations}

In Table~\ref{tab:ablation} present additional ablation experiments omitted from the main paper for space reasons.
These results confirm that our design choices help both depth and occlusion metrics.

\begin{description}
\itemsep-2pt

    \item[Naive sampling] --- Instead of our training-time sampling (described in Section 3.3. in the main paper), we sample training depths uniformly between 0.5m and 8.0m. Although this gives a slight improvement to overall IoU, it results in a drop for the harder cases of IoU Surface and IoU Boundary.
    \item[A model trained in the style of Bi3D] --- Bi3D~\cite{badki2020bi3d} proposed a stereo-matching network, which takes as input a scalar depth value. The network uses this to warp one view on to the other at the specified depth, before predicting a binary mask corresponding to the estimate of which pixels lie in front/behind the input depth value. We adapt this to work with multi-view stereo, and train this model in the spirit of Bi3D on ScanNetv2, using our backbone.
    Specifically, we warp seven source frames using the query depth plane in our reference frame and concatenate the features at each location. We then combine this with features from our image encoder as in ours.
    This method struggles to predict accurate occlusions compared to our approach. 
    We hypothesise that this is because \cite{badki2020bi3d} was designed for stereo and not for MVS.
\end{description}

\begin{table*}
    \centering
    \footnotesize
    \begin{tabular}{lccc}
        \toprule

        \textbf{Method} & IoU All$\uparrow$ & IoU Surface$\uparrow$ & IoU Boundary$\uparrow$  \\

        \midrule 
        Bi3D~\cite{badki2020bi3d} & 34.20 & 20.43 & 23.98 \\
        Ours (naive sampling) & \textbf{60.28} & 33.18 & 36.16 \\
        Ours & {60.14} & \textbf{33.29} & \textbf{36.54}
        \\
        
        \bottomrule
        
    \end{tabular}
    
    \vspace{3pt}
    \caption{\textbf{Additional ablations of binary depth handling and depth sampling.} 
    Here we compare to~\cite{badki2020bi3d} style inference and a version of our model that naively samples depth during training.  
    All models are trained equivalently on the ScanNetv2 dataset, with the same backbone network as SimpleRecon (ResNet).
    In all instances, our full model outperforms the alternatives, leading to better occlusion scores.
    }
    
    \label{tab:ablation}
\end{table*}

\subsection{Additional datasets}
Here we perform an additional quantitative evaluation on the challenging 7Scenes~\cite{shotton2013scene} and  HyperSim~\cite{hypersim} datasets. 
For the evaluation on 7Scenes, we used a model trained on ScanNetv2, while for HyperSim we use the original training/test splits.
These experiments are reported in Table~\ref{tab:7scenes_hypersim}. 
We again found that our approach outperforms regression baselines on these two datasets.

\begin{table*}
    \centering{
    \begin{tabular}{lllccccc}
        \toprule
        \textbf{Training} & \textbf{Test} & \textbf{Architecture / Method}
         & IoU All$\uparrow$ & IoU Surf.$\uparrow$ & IoU Bound.$\uparrow$ & Abs Rel$\downarrow$ \\
         \midrule
         ScanNetv2 & 7Scenes & SimpleRecon [79] & 32.42 & 16.33 & 19.90 & 0.0570 \\
         ScanNetv2 & 7Scenes & SimpleRecon [79] + Ours & \textbf{34.17} & \textbf{18.04} & \textbf{21.91} & \textbf{0.0565} \\
         \midrule
         HyperSim & HyperSim & SimpleRecon [79] & 79.42 & 52.48 & 63.22 & 0.1084 \\
         HyperSim & HyperSim & SimpleRecon [79] + Ours & \textbf{79.99} & \textbf{56.05} & \textbf{66.26} & \textbf{0.1004} \\   
        \bottomrule
    \end{tabular}
    }
    \vspace{1pt}
    \caption{\textbf{Evaluation of our method on additional datasets.} Our method outperforms a regression baseline on these two datasets.}
    \label{tab:7scenes_hypersim}
\end{table*}

\subsection{Comparison with SoTA monocular depth}
While in the main paper we focused on depth prediction from multiple frames, our method can also produce state-of-the-art (SoTA) results for monocular depth estimation. 
Here we compare with LeReS~\cite{Wei2021CVPR} a recent monocular depth prediction method.
Since LeReS uses a much larger backbone than ours (ResNeXt101 vs our ResNet18), we report experiments for a mono version of our model with a comparably large backbone (ResNeXt101). 
The results for this experiment are reported in Table~\ref{tab:monocular_comparison}. 
LeReS requires rescaling of the depth maps based on \emph{ground truth} depth at test time, which gives a significant advantage to this method. 
For a fairer comparison we report results for our method with and without rescaling. A qualitative comparison also shows the benefits of our method.

\begin{table}
    \centering{
    \resizebox{1.0\columnwidth}{!}{
        \begin{tabular}{lllccccc}
            \toprule
            \textbf{Training} & \textbf{Test} & \textbf{Architecture / Method}
             & Abs Rel $\downarrow$ & Abs Rel $\downarrow$ rescaled \\
             \midrule
             Diverse data & ScanNetv2 & LeReS~\cite{Wei2021CVPR} & -  & 0.0899 \\
             ScanNetv2 & ScanNetv2 & ResNeXt101 Mono & 0.1021  & 0.0511 \\
             ScanNetv2 & ScanNetv2 & ResNeXt101 Mono + Ours & 0.1030 &  0.0530 \\   
            \bottomrule
            \\
        \end{tabular}
    }
    \includegraphics[width=\linewidth]{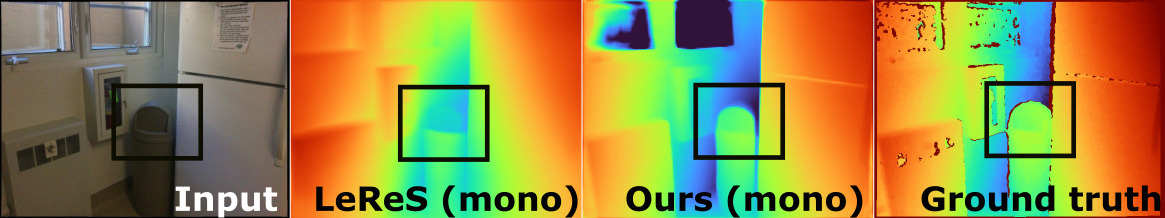}
    }
    \vspace{-10pt}
    \caption{\textbf{Comparison with state-of-the-art monocular depth.} Our method outperforms LeReS~\cite{Wei2021CVPR}, a state-of-the-art method for monocular depth estimation, when using a similar rescaling procedure at test time.}
    \label{tab:monocular_comparison}
\end{table}

\section{Additional implementation details}

\subsection{Backbone network}
Our backbone network is based on SimpleRecon~\cite{sayed2022simplerecon} which uses plane sweeping to build a cost volume and a U-Net++~\cite{zhou2018unet++} to output a final depth map. We base our experiments on the author's publicly available code.

First, a shallow feature extractor extracts features from the target, $I_\text{real}$, and source view frames. Source view features are warped to the viewpoint of the target frame at multiple hypothesis depth planes using known camera intrinsics and extrinsics. A cost volume is then created by either taking the dot product of each target and source view pair at each depth plane as in \cite{duzceker2021deepvideomvs}, or by a learned MLP \cite{sayed2022simplerecon}.
The cost volume is then passed to a cost volume encoder, together with deep image features extracted from the target frame for further refinement. This is followed by a decoder, which broadly follows the architecture of \cite{godard2019digging, watson2021multidepth, watson2019depthints}, \ie multiple convolutional layers, upsampling and skip connections from the shallow feature extractor, and the downsampling network in a U-Net++ style~\cite{zhou2018unet++}. 

For the SimpleRecon (ResNet) variant, we use a ResNet18 network for our image encoder, a dot product between source and target features in the cost volume, and a lightweight U-Net decoder from \cite{godard2019digging}. 

\subsection{Regularization around depth discontinuities}
\begin{figure}
    \centering
    \includegraphics[width=1.0\columnwidth]{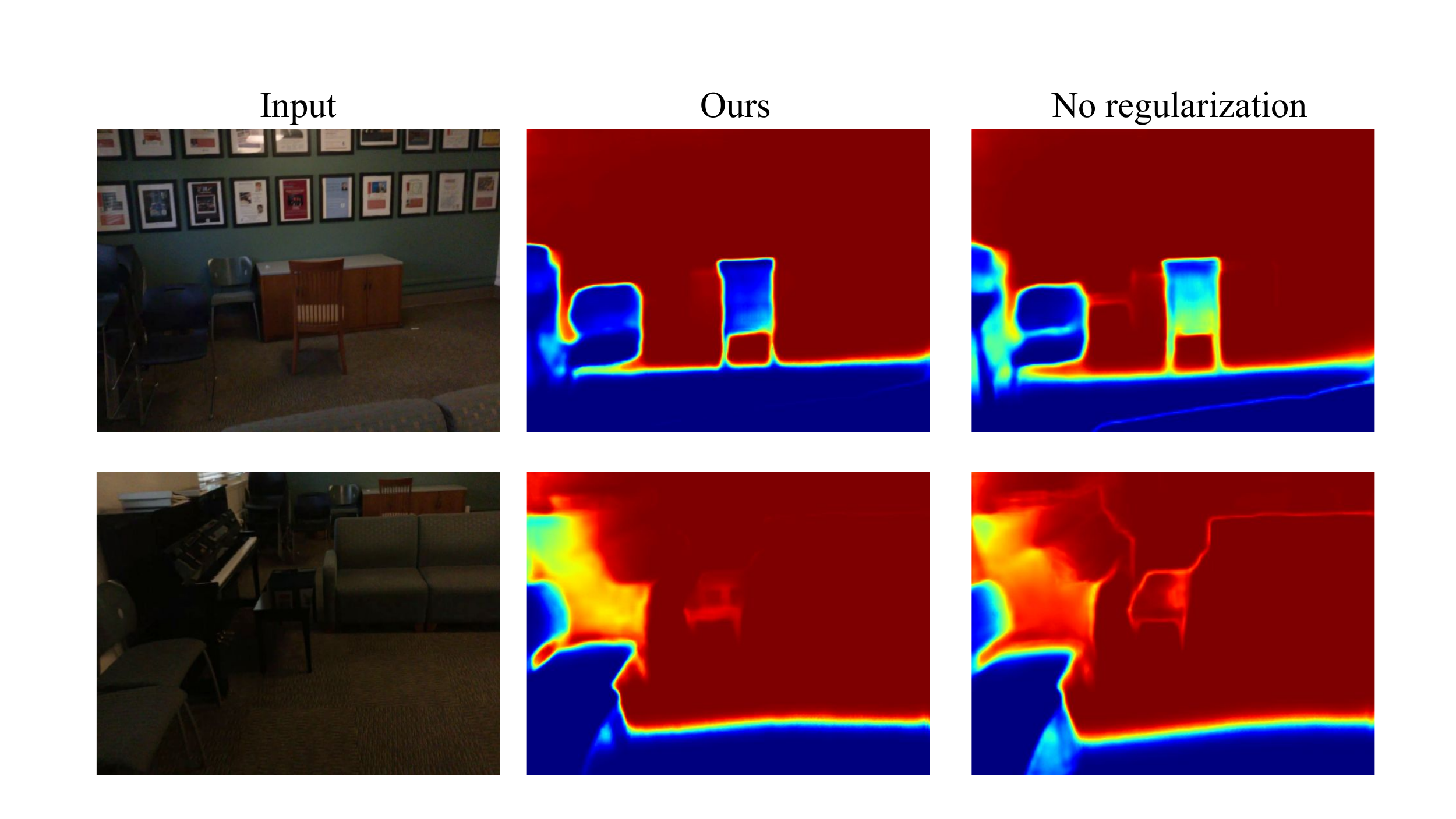}
    \caption{\textbf{With and without depth regularization.}  Example predictions for a vertical plane two meters from the camera from a model trained with and without our regularization. Blue indicates that the plane is \emph{behind} the real world geometry, and red indicates the plane is \emph{in front} of the real world geometry. Notice the white region of uncertainty in the predictions of the no regularization model on the right column. This is present both when the virtual plane is located far behind the observed scene (first row, bottom right region) and also when the virtual plane is far in front of the scene (second row, top right region).
    }    
    \label{fig:edge_reg}
    \vspace{-5pt}
\end{figure}

As discussed in Section 3.3 of the main paper, we introduce an L1 regularizer during training to discourage excessively uncertain predictions near depth discontinuities. 
Without this, predictions can have thin regions of high uncertainty which leads to a less visually pleasing final compositing result. See Figure~\ref{fig:edge_reg} for example predictions with and without regularization, for vertical planes at two meters from the camera.

To apply our regularization, we locate depth discontinuities in the ground truth depth maps using a Sobel filter~\cite{kanopoulos1988design}, and threshold to obtain an edge mask $M$. For our threshold, we use the 95th percentile of the per-image Sobel response. We subsequently apply an L1 loss during training to penalize predictions near 0.5, \ie
\begin{align}
    L_{reg} = \frac{2}{|M|} \sum_{i \in M} 0.5 - |C_i - 0.5|.
\end{align}

\subsection{Hypersim training details}

We use ScanNetv2-trained model weights from~\cite{sayed2022simplerecon} as pretrained weights for Hypersim~\cite{hypersim} regression model training.
These Hypersim-trained regression model weights %
are then used to train our Hypersim implicit depth model for occlusions. 
We removed scenes which are `broken' in the dataset from the training and validation sets, following advice from the dataset providers\footnote{\url{https://github.com/apple/ml-hypersim/issues/22}}.

\begin{figure}
    \centering
    \includegraphics[width=1\columnwidth]{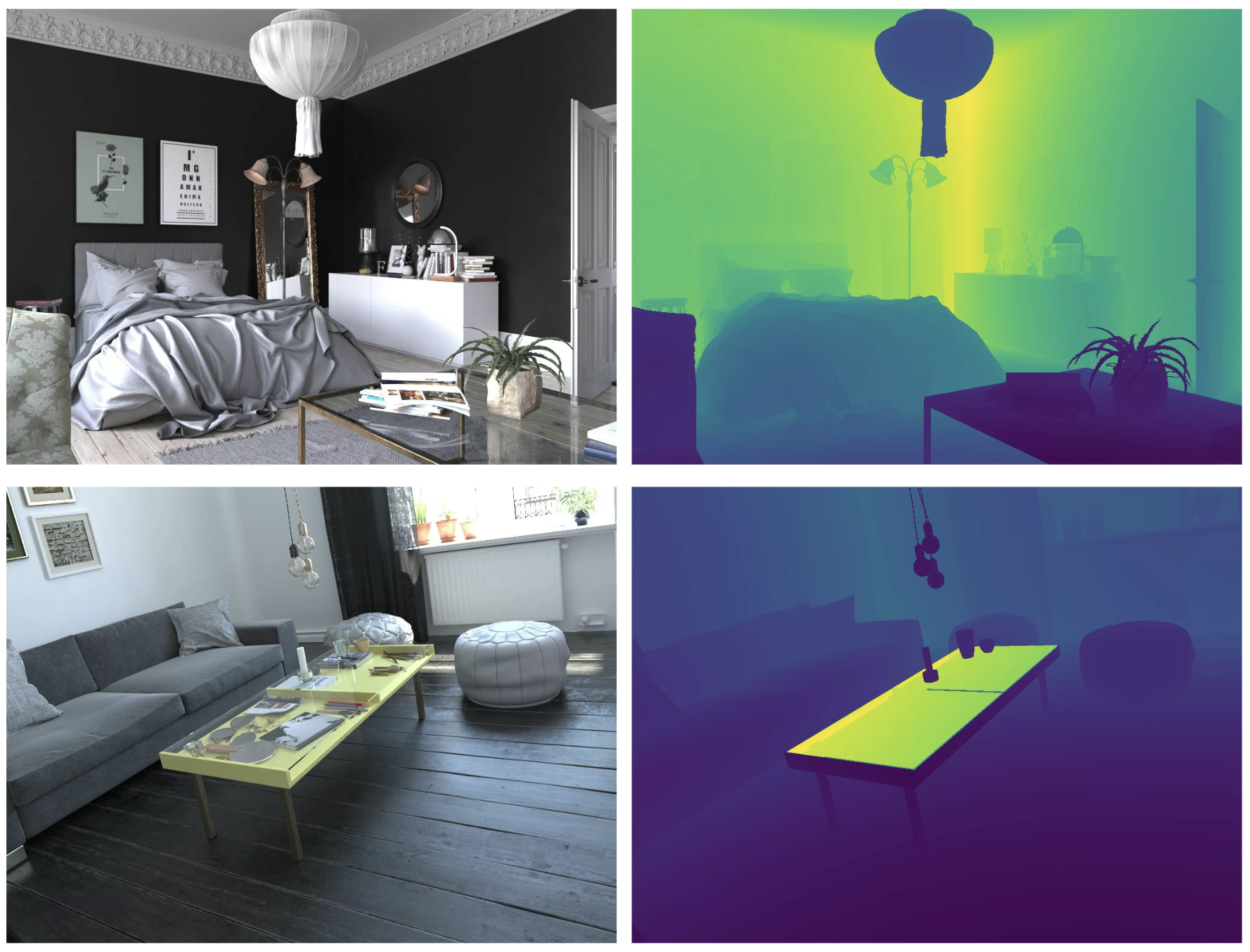}
    \caption{Example of a typical Hypersim training frame (above) and a bad frame (below) where the depth of an asset (table top) is incorrect.}
    \label{fig:hypersim-bad-depth}
    \vspace{-5pt}
\end{figure}

To make regression baselines work on Hypersim we do not use the gradient, normals and multi-view losses from~\cite{sayed2022simplerecon}.
This is due to instability we found during training, which we observed to especially occur in regions where the depth for reflective objects appears to be represent reflected depth instead of the depth at the object as shown in Figure~\ref{fig:hypersim-bad-depth}.
These large depths, occuring next to small depths, gave extremely large values for normal and gradient losses.
We attempted to solve this issue with clipping, but the issues persisted. 
We found that the Hypersim dataset gave good results even with these losses disabled.
Note that our method was not susceptible to these issues, as we train in a pixel-wise manner without gradient or normal losses. 

\begin{figure}
    \centering
    \includegraphics[width=1\columnwidth]{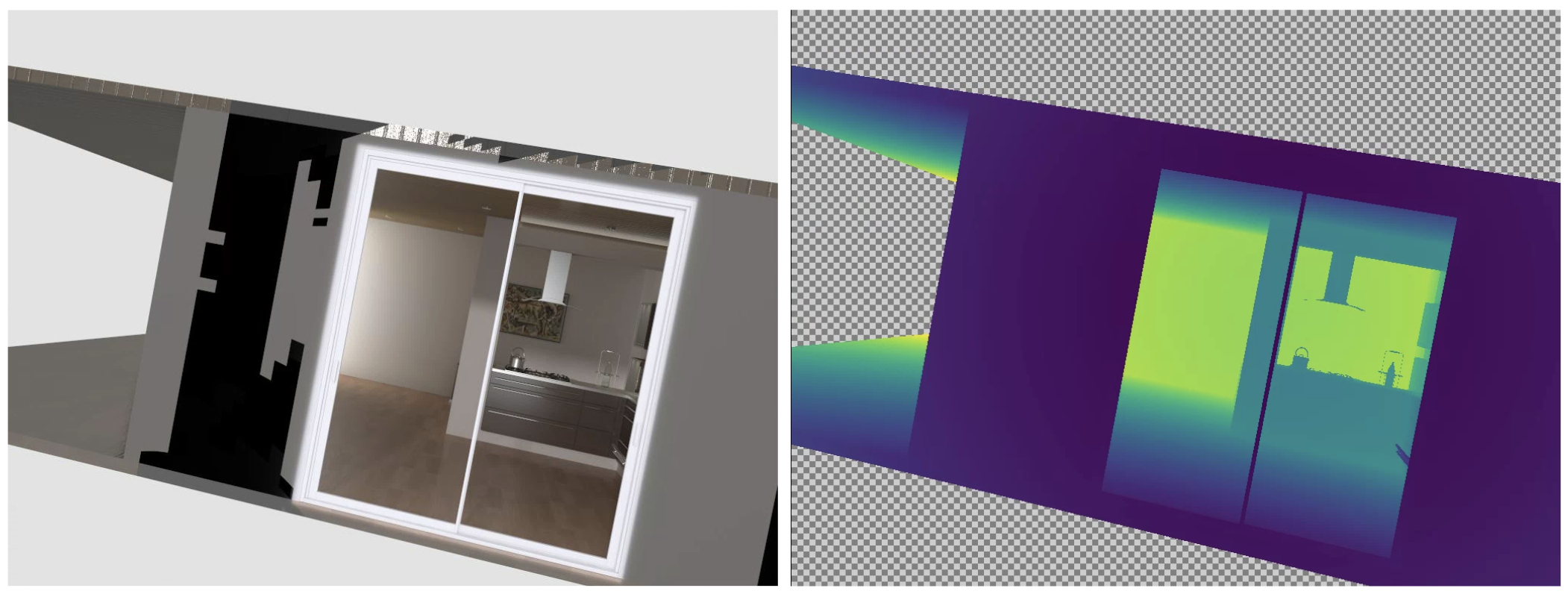}
    \caption{Example of a bad Hypersim training image that is removed by our pre-filtering.}
    \label{fig:hypersim-removed}
    \vspace{-5pt}
\end{figure}

To further help with training stability, we filtered out frames where the camera is too far away from the scene or too close to an asset.
To do this, we discard images where the most common RGB or depth pixel is greater than 30\% of the total pixel count for that image.
An example of an image removed using this filtering is shown in Figure~\ref{fig:hypersim-removed}.
Finally, after filtering and our keyframe selection step, we are left with approximately $11,500$ training frames.

To be close to ScanNetv2's depth range (where the depth bins in the cost volume range from 0.25m to 5.0m), we filter out scenes where the maximum of the mean depths for all the frames in that scene is greater than 10m. 
Due to large translations per frame, we use a maximum baseline of 2.5m for keyframe selection.
For Hypersim, due to much sharper depth edges in the dataset, we do not need to use edge regularization.

\section{Additional evaluation details}

\subsection{Plane evaluation full details}

Here we expand on the description of the occlusion task from Section~4.1 of the main paper. 
We compute the final IoU for each evaluation plane using the harmonic mean, $\text{IoU All}^d = \frac{2 \text{IoU}_+^{d}\text{IoU}_-^{d}}{\text{IoU}_+^{d} + \text{IoU}_-^{d}}$. We average the IoU for each keyframe from~\cite{duzceker2021deepvideomvs} and then for each depth plane.
For each image in the ScanNetv2 test set, we place a virtual vertical plane with a normal facing towards the camera at a fixed distance from the camera focal point. We use planes going from 1.5m up to 5m, spaced 0.5m apart, and evaluate the compositing mask predictions for each plane against the corresponding ground truth compositing mask $C_\text{true}$. 

To focus evaluation on regions of difficulty, we also evaluate IoU around the ground truth edge boundary (\ie around the Trimap), as advocated by \cite{chen2017deeplab}, referred to as \textit{IoU Boundary}, as well as near the ground truth geometry surface, referred to as \textit{IoU Surface}. 
For IoU Boundary, we evaluate all pixels which lie within seven pixels of the ground truth edge boundary.
For IoU Surface, we evaluate any samples for which the rendered asset depth is within 5\% of the ground truth depth.

\subsection{Temporal consistency evaluation details}

For our temporal consistency evaluation from Section 4.3 in the main paper, we use all 100 scenes from the ScanNetv2 test set.
For each scene, we take the first 120 frames which are then divided into 8 equal sub-sequences of 15 frames. 
For each of these sub-sequences, the vertical evaluation plane is placed at a fixed location directly in front of the camera location of the first frame in the sub-sequence.
The plane is placed at a depth $d_\text{eval}$ from the camera, where $d_\text{eval}$ is computed as the 75th percentile of the ground truth depths values of this first frame in the sub-sequence. 
This is done to ensure we always have a valid plane in front of the camera for evaluation.
The first 2 frames are not used in the metric computation, \ie only 13 frames per sub-sequence are used. The 2 frames at the start of the sub-sequences are considered  ``warmup'', which is important to obtain representative results for multi-view methods. %

\subsection{Fast densification baseline details}

For evaluating the baseline (`Depth Dens. w/ Lidar') from~\cite{holynski2018fast}, we used the Python code provided by the authors\footnote{\url{https://github.com/facebookresearch/AR-Depth}}.
Their code requires sparse 2D points on each input image with associated per-point depth.
These are assumed to be collected from a SLAM or SfM system.
For our sequences captured from the Apple iPhone, we found the points returned were far too sparse, \ie each frame had only between 100 and 200 points returned, while the demo sequence from~\cite{holynski2018fast} had around an order of magnitude more points per frame.

So instead, we randomly sampled 2,000 points in image space.
To help improve temporal stability, for each sequence we used the same fixed 2,000 points for each image.
We also experimented with using a KLT tracker to provide points, but these tended to be excessively clustered on textured regions and led to worse performance. 
For each of our 2,000 points, we used the depth map from Apple Lidar to give each point a depth value.
This implementation of the \cite{holynski2018fast} baseline is therefore an improved version as it has access to Lidar, and highlights their performance in a best-case scenario.

\begin{table*}[t]
    \begin{center}

    \setlength{\tabcolsep}{0.3em} %

    \begin{tabular}{lcccccccc}
        \toprule
        & Abs Diff$\downarrow$  & Abs Rel$\downarrow$ & Sq Rel$\downarrow$ & RMSE$\downarrow$ & logRMSE$\downarrow$ & $\delta < 1.05\uparrow$ & $\delta < 1.10\uparrow$ &  $\delta < 1.25\uparrow$  \\
        \midrule
        DPSNet (FT)~\cite{im2019dpsnet} & .1552 & .0795 & .0299 & .2307 & .1102 & 49.36 & 73.51 & 93.27 \\
        MVDepthNet (FT)~\cite{wang2018mvdepthnet}  & .1648 & .0848 & .0343 & .2446 & .1162 & 46.71 & 71.92 & 92.77 \\
        DELTAS~\cite{sinha2020deltas} & .1497 & .0786 & .0276 & .2210 & .1079 & 48.64 & 73.64 & 93.78\\
        GPMVS (FT)~\cite{hou2019multi}  & .1494 & .0757 & .0292 & .2287 & .1086 & 51.04 & 75.65 & 93.96\\
        DeepVideoMVS, pairnet\textdagger*~\cite{duzceker2021deepvideomvs} & .1431 & .0712 & .0253 & .2152 & .0999 & 51.92 & 77.24 & 94.99 \\
        DeepVideoMVS, fusion\textdagger*~\cite{duzceker2021deepvideomvs} &
         .1186 & .0583 & .0190 & .1879 & .0868 & 60.20 & 83.66 & 96.76 \\
        SimpleRecon (ResNet) & .0978 & .0487 & .0151 & .1617 & .0752 & 69.52 & 88.13 & 97.46 \\
        SimpleRecon~\cite{sayed2022simplerecon} & .0871 & \textbf{.0429} & .0125 & .1460 & .0672 & \textbf{74.01} & \textbf{90.75} & \textbf{98.08} \\
        \midrule
        SimpleRecon (ResNet) + \textbf{Ours} & .0988 & .0498 & .0149 & .1595 & .0743 & 68.52 & 87.53 & 97.42 \\
        SimpleRecon + \textbf{Ours} & \textbf{.0862} & .0436 & \textbf{.0123} & \textbf{.1426} & \textbf{.0666} & 73.74 & 90.54 & 98.06 \\

        \bottomrule
    \end{tabular}

    \end{center}
    \vspace{-4pt}
    \caption{\textbf{Additional depth metrics for ScanNetv2.} 
    Here we expand on Table 1 from the main paper by adding the full set of standard depth metrics used in existing work, \eg \cite{sayed2022simplerecon}.
     \textdagger~two measurement frames. *~trained on 90/10 split.}
    \label{table:depth_results_scannet}
    \vspace{-5pt}
\end{table*}

\subsection{Improving the regression baselines}
\begin{figure}[h]
    \centering
     \begin{tabular}{cc}
        \textbf{\footnotesize Without blended masks} &
        \textbf{\footnotesize With blended masks} \\
       \includegraphics[width=0.43\columnwidth]{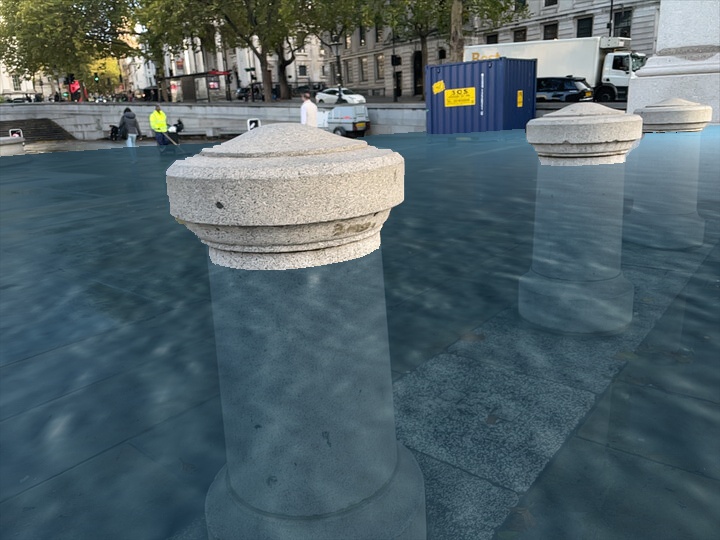} & 
       \includegraphics[width=0.43\columnwidth]{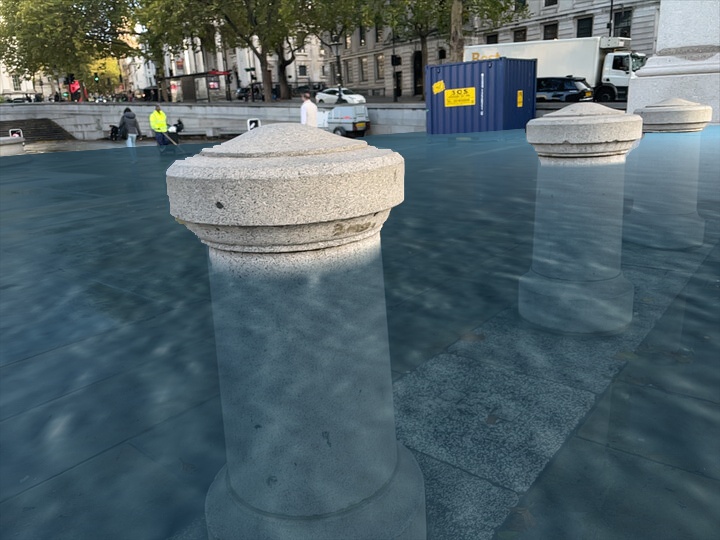}
    \end{tabular}
    
    \caption{A Hypersim-trained regression baseline without mask blending (left) and with mask blending (right). Adding the blended mask makes intersections of real and virtual geometry in the baseline systems more believable. Best viewed zoomed in. 
    }
    \label{fig:blended_mask}
    \vspace{-5pt}
\end{figure}

When making the occlusion mask with our regression baselines, we enhance the qualitative results through the use of \emph{blended masks}.
This improves the visual quality of this baseline approach.
When the predicted depth is near to the virtual depth, we create a blended mask.
Specifically, we form a mask as follows,
\begin{align}
    C &= \text{clamp} \left( \frac{D_\text{real} - D_\text{virtual} + b }{b}, 0.0, 1.0 \right),
\end{align}
where $b$ is the size of the blending band and $\text{clamp}(x, \min, \max)$ clamps the input $x$ at $\min$ and $\max$.
This expression results in a mask that linearly interpolates between $0$ and $1$ as the real depth approaches, and then goes behind the virtual depth.
We found $b=0.2m$ made the regression baselines look good.
A comparison of a regression baseline method with and without this blended mask is shown in Figure~\ref{fig:blended_mask}.

\begin{figure}[h]
    \centering
    \includegraphics[width=\columnwidth]{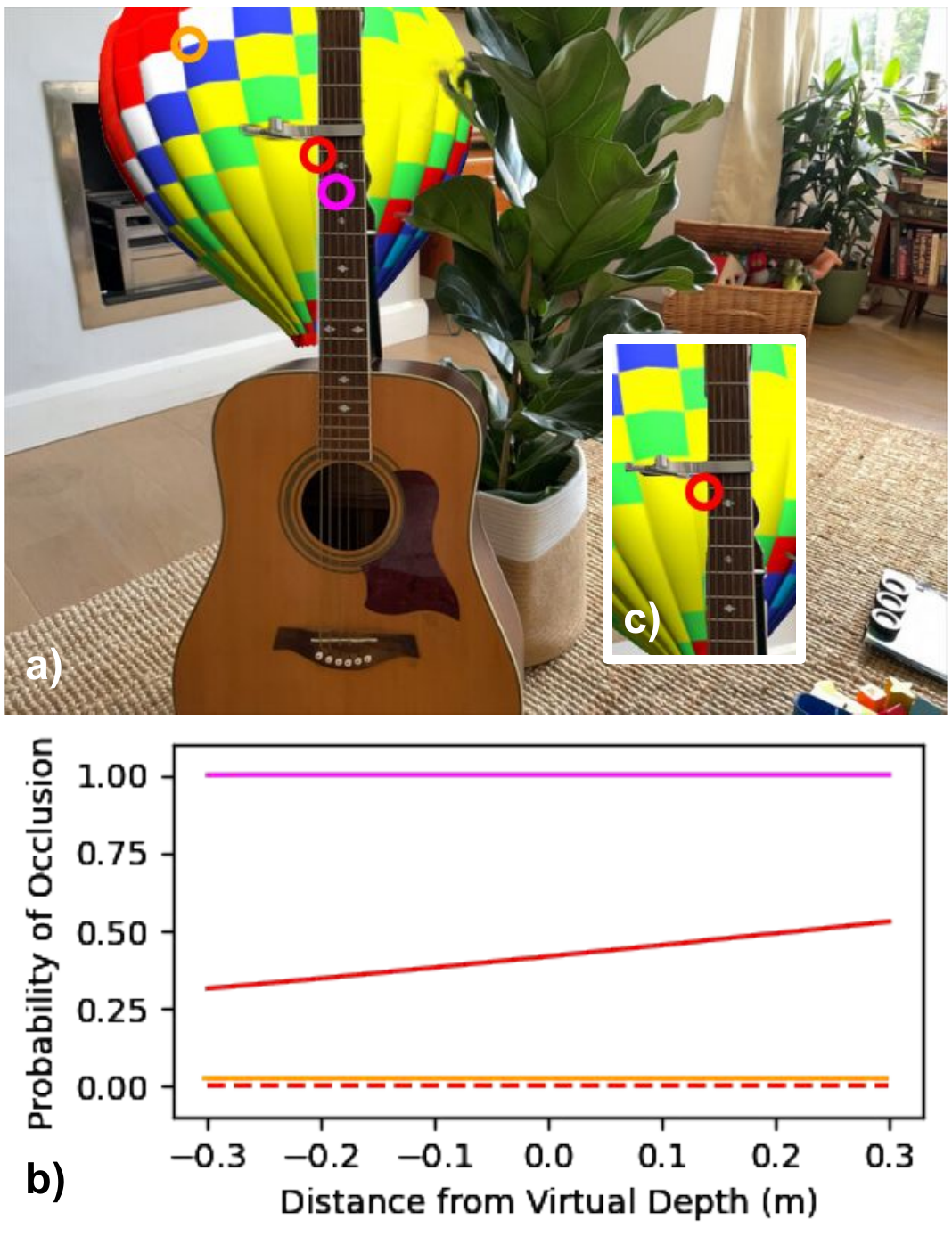}
    \caption{Visualizing the probability of the asset's occlusion given its virtual depth. a) Output from our method with points whose rays are plotted marked. b) Plot of the probability of occlusion for each marker within a range of depths from the virtual asset. c) A regression baseline output for the same frame. See red marker: Our method outputs softer probabilities around edges (solid red line in plot), whereas the regression baseline has to output a hard decision at each depth location (dashed red line), leading to hard incorrect boundaries around occlusions. 
    }
    \label{fig:ray_prob}
    \vspace{-5pt}
\end{figure}

\subsection{Complete depth results}

In Table~\ref{table:depth_results_scannet} we show the depth results with additional depth evaluation metrics that were omitted from Table 1  in the main paper for space reasons.
We notice again our model is the state-of-the-art in these metrics, matching or beating prior work across each of the eight categories.

\subsection{Probability visualization}

Figure~\ref{fig:ray_prob} shows the probability predicted by our model for some specific pixels in a test image, for different values of virtual depth.
See the caption for more details.

\end{document}